\documentclass[runningheads]{llncs}

\usepackage{eccv}

\usepackage{eccvabbrv}

\usepackage{graphicx}
\usepackage{booktabs}
\usepackage{multirow}
\usepackage{tikz}
\usepackage{pgfplots}
\usepackage{bm}
\usepackage{xcolor}
\pgfplotsset{compat=1.18}
\usetikzlibrary{positioning,arrows.meta,calc,fit,backgrounds,decorations.pathreplacing,shapes.geometric}
\usepackage{pifont}
\usepackage{tcolorbox}

\usepackage[accsupp]{axessibility}  %

\usepackage[pagebackref,breaklinks,colorlinks,citecolor=eccvblue]{hyperref}
\usepackage{hyperref}

\usepackage{orcidlink}

\newread\imgstream
\immediate\openin\imgstream=imagedata.in
\makeatletter
\def\new@kvginclip#1{}
\def\new@kvgintrim#1{}
\let\old@kvginclip\KV@Gin@clip
\let\old@kvgintrim\KV@Gin@trim
\let\oldincludegraphics\includegraphics
\providecommand{\includegraphics}{}
\renewcommand{\includegraphics}[2][]{%
  \immediate\read\imgstream to \src
  \immediate\read\imgstream to \removecrop
  \ifnum\removecrop=1
      \let\KV@Gin@clip\new@kvginclip
      \let\KV@Gin@trim\new@kvgintrim
  \fi
  \oldincludegraphics[#1]{\src}%
  \let\KV@Gin@clip\old@kvginclip
  \let\KV@Gin@trim\old@kvgintrim}
\makeatother

\begin{document}

\title{GeoFlow: Enforcing Implicit Geometric Consistency in Video Generation}

\titlerunning{GeoFlow: Enforcing Implicit Geometric Consistency in Video Generation}

\author{Jan Ackermann\inst{1} \and
Shengqu Cai\inst{1}\textsuperscript{*} \and
Boyang Deng\inst{1}\textsuperscript{*} \and
Zhengfei Kuang\inst{1}\textsuperscript{*}\\
Songyou Peng\inst{2} \and
Gordon Wetzstein\inst{1}}

\authorrunning{J.~Ackermann et al.}

\institute{Stanford University \and Google DeepMind}

\maketitle
\begingroup
\renewcommand{\thefootnote}{\fnsymbol{footnote}}
\footnotetext[1]{Equal contribution.}
\endgroup
\begin{abstract}
Generating geometrically consistent videos remains an open challenge: text-to-video diffusion models trained on web-scale data treat geometry only implicitly, leading to object deformation, texture drift, and non-rigid backgrounds under camera motion. Existing solutions either improve consistency as a byproduct, apply only to static scenes or realign the latent space of the model completely. We introduce a geometry-consistency reward that directly measures whether motion in a generated video is compatible with a coherent scene. Our key insight is that in physically consistent videos, background motion should be explainable by rigid camera-induced flow, while independently moving objects should preserve appearance identity along motion trajectories. We operationalize this using optical flow, depth--pose predictions, and feature-based correspondence to separate rigid and dynamic regions and evaluate their respective consistency. Integrating this reward with reinforcement fine-tuning transforms geometric consistency from an emergent property into an explicit optimization objective for video generators. The approach is model agnostic and applies to diverse dynamic scenes containing both camera and object motion. Experiments show substantial reductions in temporal geometric artifacts over strong baselines while preserving perceptual quality. Code and model weights are available at \url{https://geometryflow.github.io/}.
\end{abstract}
\section{Introduction}

High-quality 3D and 4D visual content is increasingly central to emerging technologies, e.g. AR/VR, simulation, and robotics. However, producing temporally and geometrically consistent assets remains challenging. While large-scale 3D/4D datasets are scarce and costly to curate, web-scale 2D video is abundant and drives the most capable generative models~\cite{rombach2022high, batifol2025flux, hong2022cogvideo}. A key opportunity is therefore to learn spatially and temporally coherent generation directly from video. 

Despite rapid progress in text-to-video diffusion models, current systems largely treat geometry implicitly. As a result, generated videos often exhibit temporally inconsistent structure: objects deform over time, textures drift across surfaces, and backgrounds behave non-rigidly under camera motion. These artifacts arise because standard likelihood-based objectives do not explicitly enforce that pixel trajectories correspond to a coherent 3D scene. Bridging this gap requires training signals that reward geometric consistency over time.

Existing methods address this need from complementary directions but share important limitations. 3D foundation models learn powerful spatial priors within explicit representations~\cite{yang2019pointflow, chen2024meshxl,chen20253dtopia}, yet remain constrained by limited 3D/4D data. Camera-controlled video models~\cite{he2024cameractrl, bahmani2024vd3d} inject geometry through pose conditioning or explicit 3D buffers, but are still optimized for conditional likelihood rather than geometric correctness, leading to fragile temporal coherence under complex motion. Post-hoc alignment methods~\cite{wu2025geometry, xie2024carve3d} adapt pretrained generators using geometric supervision but often rely on object-level priors, curated 3D teachers, or reconstruction pipelines, which limit generality and scalability.
\begin{figure}[t!]
    \centering
    \includegraphics[width=\linewidth]{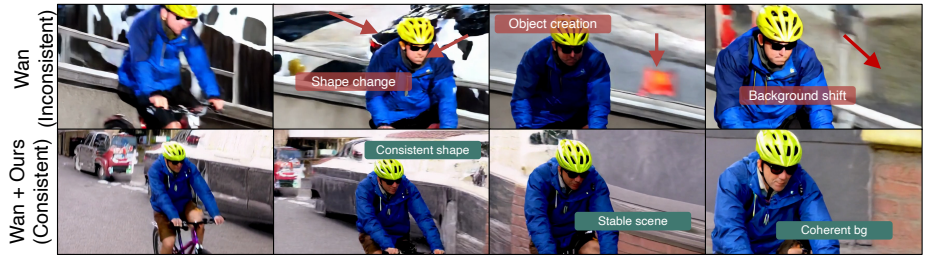}
    \caption{\textbf{Improving the geometric consistency of generated videos.} Video models still struggle to produce geometrically consistent videos. The top row shows an example generated by Wan2.1-1.3B~\cite{wan2025wan}, which exhibits object deformation as well as background and scene shifts. The bottom row shows the same video generated after training with our method, where these failures are no longer visible.}
    \label{fig:teaser}
\end{figure}

We propose directly optimizing geometric consistency in video generation via a \emph{geometry-consistency reward}. Our key insight is that temporal geometric coherence can be effectively approximated through motion analysis. In a consistent scene, background motion induced by camera movement can largely be explained by a rigid 3D transformation, while independently moving objects should preserve their appearance along their motion trajectories. 
Accordingly, we decouple evaluation into low-level structural integrity and high-level semantic alignment, allowing the reward to handle highly dynamic scenes while avoiding a strict rigid-world assumption. We materialize this principle by using optical flow and geometry predictions from a pretrained model to disentangle rigid from dynamic regions, allowing us to measure their respective consistency. This produces a dense, self-supervised reward that quantifies whether observed motion is compatible with a coherent scene.

We integrate this reward with reinforcement fine-tuning (Flow-GRPO \cite{liu2025flow}) to align pretrained text-to-video generators toward temporally consistent geometry while preserving visual fidelity. Crucially, the reward depends only on the generated video and pretrained geometric predictors, requiring no 3D/4D ground truth, teacher reconstructions, or pose conditioning. Because the signal is derived from generic motion and geometric cues, it applies broadly across diverse scenes containing both camera and object motion.
In practice, our alignment substantially reduces temporal artifacts such as object deformation, texture drift, and background wobble across challenging motion regimes, while maintaining perceptual quality and diversity.
Overall, our main contributions are as follows:
\begin{itemize}
    \item We introduce a geometry-consistency reward for text-to-video generation that measures whether pixel motion is explainable by a coherent 3D scene with rigid background and persistent moving objects.
    \item We adapt Flow-GRPO~\cite{liu2025flow} for high-resolution video generation, demonstrating that targeted modifications, specifically synchronized noise initialization and truncated backpropagation, make RL fine-tuning computationally viable and stable for complex geometric alignment.
    \item We demonstrate that RL fine-tuning with our reward significantly improves the temporal and geometric consistency of pretrained video diffusion models, thereby reducing deformation and motion artifacts in highly dynamic environments, without degrading visual fidelity.
\end{itemize}
\section{Related Work}

\paragraph{Video Generation.}
Recent advances in diffusion models have enabled high-quality text-conditioned video generation from large-scale web data~\cite{hong2022cogvideo, yang2024cogvideox, wan2025wan, kong2024hunyuanvideo, genmo2024mochi, videoworldsimulators2024, wiedemer2025video}. To improve spatial coherence and controllability, a common strategy is to condition generation on camera motion or scene trajectories. Camera-controlled video diffusion models incorporate pose or trajectory signals during denoising to encourage 3D-consistent motion~\cite{he2024cameractrl, zheng2024cami2v, bahmani2024vd3d, van2024generative}. Related approaches generate world-consistent videos along specified camera paths or jointly model scene dynamics and viewpoint~\cite{kuang2024collaborative, bai2024syncammaster}. Building on these structural foundations, subsequent work improves long-horizon generation through memory mechanisms and autoregressive frameworks~\cite{wu2025video, xiao2025worldmem, po2025long, song2025history, po2025bagger, zhang2025frame, cai2025mixture, chen2024diffusion}, as well as broader world-model formulations~\cite{bruce2024genie}.
Despite these advances, standard likelihood-based training does not explicitly enforce geometric consistency over time, leading to artifacts such as object deformation, texture drift, and non-rigid backgrounds under camera motion.
Our work addresses this gap by introducing an explicit geometry-consistency objective for video generation. 

\paragraph{3D-Aware Generative Models.}
As an alternative to relying purely on 2D priors and pose conditioning, another line of work induces geometry through explicit 3D or multi-view representations. Multi-view diffusion models generate consistent view sets from a single image or prompt~\cite{liu2023zero123, shi2023zero123pp, liu2023syncdreamer, shi2023mvdream, li2024era3d, cao2025mvgenmaster, yu2024viewcrafter, gao2024cat3d, wu2025cat4d, chou2025generating}. Earlier 3D-aware generative models incorporated radiance-field or meshes to enforce viewpoint consistency~\cite{chan2021pi, chan2022efficient, chan2023generative, gao2022get3d, shue20233d}. More recently, 3D foundation models learn strong priors for reconstruction or rendering from limited 3D/4D observations~\cite{hong2023lrm, tochilkin2024triposr, xu2024grm, tang2024lgm, xu2025depthsplat, chen2024mvsplat}. Although these approaches provide explicit geometric structure, they are constrained by representation-specific training or scarce 3D supervision and primarily address static or quasi-static scenes.

\paragraph{Diffusion Alignment for Geometric Consistency.}
Alignment methods refine pretrained diffusion models toward downstream objectives beyond standard likelihood. Preference-based and reinforcement-learning approaches, such as DPO~\cite{rafailov2023dpo}, DDPO~\cite{wallace2024diffusiondpo}, and GRPO~\cite{shao2024grpo}, enable post-training optimization of diffusion policies~\cite{wallace2024diffusiondpo, black2023training, liang2024richhf, yuan2024spindiffusion}. Extending these principles to video generation, Flow-GRPO~\cite{liu2025flow} provides a stable RL framework for flow-based models, with further training insights from methods like Dance and MixGRPO~\cite{xue2025dancegrpo, li2025mixgrpo}.
Building on these optimization frameworks, several works extend alignment specifically toward geometric objectives. For instance, Carve3D~\cite{xie2024carve3d} applies RL fine-tuning to multi-view diffusion using reconstruction consistency, while concurrent works~\cite{kupyn2025epipolar,bai2025geovideo,du2026videogpadistillinggeometrypriors} apply geometric constraints on static videos, and IC-World~\cite{wu2025ic} aligns two views of the same scene. Other approaches inject geometric priors or perform post-hoc refinement through adapters, guided sampling, consistency modules, or representation constraints~\cite{chen2024-3dadapter, seo2023let, edelstein2025sharpit, yu2024representation, wu2025geometry, bengtson2025geometric, gu2025geco,zhang2025videorepa}. However, they often rely on strong teacher models, object-level priors, or quasi-static assumptions, which limit their generality.
We introduce a geometry-consistency reward derived directly from motion and geometry predictions to overcome these limitations. Rather than acting as an imitated rigid teacher, our reward serves solely as an evaluation signal. By combining this explicit signal with RL fine-tuning via Flow-GRPO~\cite{liu2025flow}, we can align text-to-video generators toward temporally consistent geometry in fully dynamic scenes, entirely without 3D supervision.
\section{Method}

\paragraph{Problem Setup.}
We consider text-to-video generation with geometric consistency.  
Given a text prompt $\mathbf{c}$, a video generator produces a sequence of $T$ frames
\[
\mathbf{V} = \{\bm{I}_\tau\}_{\tau=1}^{T}, \qquad \bm{I}_\tau \in \mathbb{R}^{H \times W \times 3}.
\]
We assume that each frame depicts a scene observed from an unknown camera pose and possibly containing independently moving objects.  
Let $X^*$ denote the (unknown) underlying dynamic 3D scene consisting of a rigid background and moving foreground objects.  
For any visible 3D point $x \in X^*$ and frame $\tau$, its image projection satisfies
\[
\tilde{x}_\tau \sim \bm{P}_\tau x,
\]
where $\bm{P}_\tau = \bm{K}_\tau \bm{E}_\tau$ is the camera projection matrix at time $\tau$.

A temporally consistent video is one in which pixels correspond to coherent trajectories induced by camera and object motion.  
Ideally, generator parameters $\theta$ should minimize the expected reprojection error over all 3D points and frames:
\begin{equation}
\theta^* =
\arg\min_\theta
\mathbb{E}_{\tau}
\left[
\frac{1}{|X^*|}
\sum_{x \in X^*}
\|\hat{\mathbf{u}}_\tau(x;\theta) - \bm{P}_\tau x\|_2^2
\right],
\label{eq:reproj_video}
\end{equation}
where $\hat{\bm{u}}_\tau(x;\theta)$ denotes the pixel corresponding to $x$ in generated frame $\bm{I}_\tau$.  
Since $X^*$ and $\bm{P}_\tau$ are unknown, this objective is intractable.  
We therefore introduce a geometry-consistency reward $R(\theta)$ that serves as a surrogate:
\begin{equation}
R(\theta) \;\propto\; -\,\mathcal{E}_{\text{reproj}}(\theta),
\qquad
\theta^* = \arg\max_\theta R(\theta).
\label{eq:reward_equiv_video}
\end{equation}
This casts consistent video generation as a reward maximization problem.

\begin{figure*}[t]
    \centering
    \includegraphics[width=\textwidth]{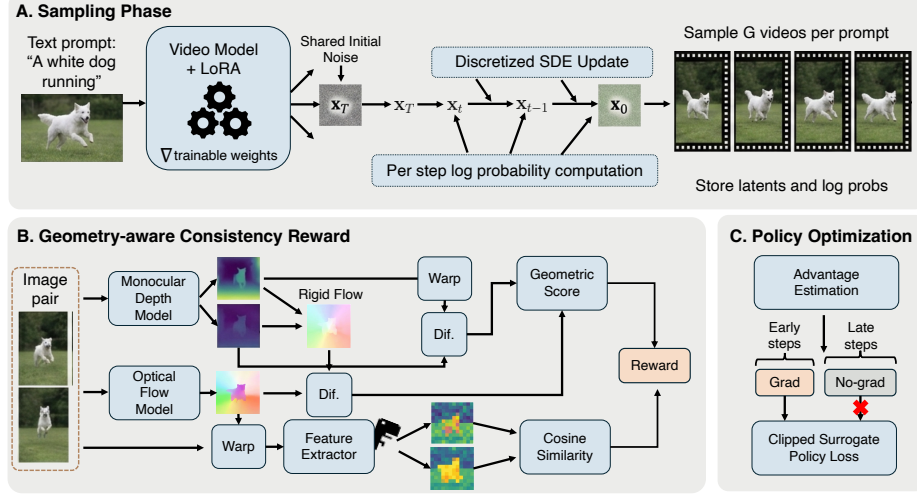}
    \caption{\textbf{GeoFlow method overview.}\textbf{(A)} A video diffusion model $\pi_\theta$ generates $G$ candidate videos from text prompts using the same initial noise. 
\textbf{(B)} A monocular depth model predicts depth maps, while an optical flow model estimates the flow between two frames. Rigid flow is derived from the predicted depth and compared with the estimated optical flow. The resulting residual flow is combined with the discrepancy between the predicted depth and the flow-warped depth to compute the \textit{Geometric Score}. In addition, semantic scene consistency is evaluated by comparing the semantic features of the second frame with those of the flow-warped first frame. 
\textbf{(C)} GRPO computes group-normalized advantages and updates $\theta$ using the clipped surrogate objective applied only to the first $M$ timesteps.}
    \label{fig:method}
\end{figure*}

\paragraph{Method Overview.}
\looseness = -1
We design a geometry-consistency reward that measures whether the motion between frames is explainable by a coherent 3D scene.  
Our key insight is that rigid background motion induced by camera motion should be predictable from geometry, whereas independently moving objects should exhibit appearance consistency along their motion trajectories. 
This is implemented via two complementary signals:
(i) rigid motion consistency between observed optical flow and geometry-induced flow, and (ii) dynamic identity consistency measured in a learned feature space along flow trajectories.
Instead of relying on hard segmentation masks to separate these regions, we formulate a joint continuous quality score where structural geometric consistency naturally acts as a soft weight. The complementary semantic reward evaluates appearance globally, natively accommodating independent motion without requiring explicit tracking.
The resulting reward can be integrated with any policy-gradient fine-tuning method; we instantiate it with our adaptation of Flow-GRPO~\cite{liu2025flow}. It is also efficient to compute: on a single NVIDIA H100, it processes videos at 8 FPS---significantly faster than videos can be sampled on the same hardware. \cref{fig:method} shows an overview of our method.

\subsection{Geometry-Consistency Reward}
\label{sec:geometry-reward}
We propose a geometry-aware reward formulation designed to explicitly quantify the spatio-temporal consistency of generated videos. By decoupling the evaluation into low-level structural integrity and high-level semantic alignment, our reward reliably scores highly dynamic scenes without strictly enforcing a rigid-world assumption. 

\paragraph{Monocular Priors and Flow Estimation.}
Given a generated video, we sample consecutive frame pairs $(\bm{I}_\tau, \bm{I}_{\tau+1})$. We extract foundational geometric representations using a robust monocular depth prior (e.g., Depth Anything 3~\cite{lin2025depth}), which yields dense metric depth maps $\bm{D}_\tau$, camera intrinsics $\bm{K}_\tau$, and world-to-camera extrinsics $\bm{E}_\tau$. Concurrently, we estimate the dense observed optical flow $\bm{F}_{\mathrm{pred}}$ between frames using an off-the-shelf flow network (e.g., WAFT~\cite{wang2025waft}). Finally, to capture deep visual semantics, we utilize a pre-trained vision transformer (e.g., DINOv2~\cite{oquab2023dinov2}) to extract patch-level feature embeddings $\phi(I_\tau)$.

\paragraph{Geometric Structural Consistency.}
To evaluate 3D coherence, we derive a theoretical rigid flow field, $\bm{F}_{\mathrm{rig}}$, which represents the apparent pixel motion induced purely by the camera's movement through a static scene. We compute the relative camera pose $\bm{T}_{\tau \to \tau+1} = \bm{E}_{\tau+1} \bm{E}_\tau^{-1}$. For each source pixel $\mathbf{u} = (u, v)$, we unproject it to 3D using $\bm{D}_\tau(\mathbf{u})$ and $\bm{K}_\tau$, apply the transformation $\bm{T}_{\tau \to \tau+1}$, and reproject it to the target viewpoint to compute the spatial displacement $\bm{F}_{\mathrm{rig}}(\mathbf{u})$. 

We define the endpoint error (EPE) between the observed flow and the rigid flow as $E_{\mathrm{epe}}(\mathbf{u}) = \|\bm{F}_{\mathrm{pred}}(\mathbf{u}) - \bm{F}_{\mathrm{rig}}(\mathbf{u})\|_2$. To ensure robustness against large camera motions, we compute a normalized EPE:
\begin{equation}
\bar{E}_{\mathrm{epe}}(\mathbf{u}) = \frac{E_{\mathrm{epe}}(\mathbf{u})}{\|\bm{F}_{\mathrm{pred}}(\mathbf{u})\|_2 + \|\bm{F}_{\mathrm{rig}}(\mathbf{u})\|_2 + \bm{\epsilon}_{\mathrm{num}}} ,
\label{eq:epe_norm}
\end{equation}
where $\bm{\epsilon}_{\mathrm{num}}$ (e.g., $1$px) stabilizes the penalty in perfectly static regions. Additionally, we warp the source depth map to the target viewpoint (yielding $\tilde{\bm{D}}_{\tau \to \tau+1}$) and compute the relative depth error against the predicted target depth $\bm{D}_{\tau+1}$:
\begin{equation}
E_{\mathrm{depth}}(\mathbf{u}) = \frac{|\tilde{\bm{D}}_{\tau \to \tau+1}(\mathbf{u}) - \bm{D}_{\tau+1}(\mathbf{u})|}{\bm{D}_{\tau+1}(\mathbf{u}) + \bm{\epsilon}_{\mathrm{num}}} .
\label{eq:depth_err}
\end{equation}
The joint geometric quality score per pixel is formulated as the product of the bounded constraints, allowing the rigid flow score to act as a soft weight on the depth error:
\begin{equation}
Q_{\mathrm{geo}}(\mathbf{u}) = \big(1 - \min(\bar{E}_{\mathrm{epe}}(\mathbf{u}), 1)\big) \cdot \big(1 - \min(E_{\mathrm{depth}}(\mathbf{u}), 1)\big) .
\label{eq:geo_quality}
\end{equation}
The geometric reward component, $R_{\mathrm{geo}}$, is the normalized average across all valid pixels (masked by depth availability and optional model confidence gating), shifted to a $[-1, 0]$ range via $R_{\mathrm{geo}} = \frac{1}{|\Omega|} \sum_{\mathbf{u} \in \Omega} Q_{\mathrm{geo}}(\mathbf{u}) - 1$, where $\Omega$ is the set of valid pixels.

\paragraph{Semantic Dynamic Consistency.}
While $R_{\mathrm{geo}}$ strictly enforces static background geometry, moving objects will naturally violate rigid constraints. To accommodate independent object motion, we evaluate semantic identity preservation along the actual motion trajectories. Using the predicted flow $\bm{F}_{\mathrm{pred}}$, we differentiably warp the source image to synthesize $\tilde{\bm{I}}_{\tau \to \tau+1}$. We then extract deep features from this warped image, yielding $\phi(\tilde{\bm{I}}_{\tau \to \tau+1})$. 
We quantify consistency via the patch-wise cosine distance:
\begin{equation}
R_{\mathrm{dino}} = - \frac{ \sum_{\mathbf{p}} \bm{W}(\mathbf{p}) \big[ 1 - \cos\!\big( \phi(\tilde{\bm{I}}_{\tau \to \tau+1})(\mathbf{p}), \phi(\bm{I}_{\tau+1})(\mathbf{p}) \big) \big] }{ \sum_{\mathbf{p}} \bm{W}(\mathbf{p}) } ,
\label{eq:Rdino}
\end{equation}
where $\mathbf{p}$ indexes the spatial patches, and $\bm{W}(\mathbf{p})$ is a binary mask excluding out-of-bounds warped pixels. This formulation globally penalizes identity drift, inherently accommodating both static backgrounds and dynamic foreground objects.

\paragraph{Composite Reward.}
The final consistency reward for a frame pair is a linear combination of structural and semantic alignment:
\begin{equation}
R_\tau = \lambda \cdot \, R_{\mathrm{geo}} + (1-\lambda)\cdot\, R_{\mathrm{dino}},\  \lambda\in[0,1] .
\label{eq:Rpair}
\end{equation}
The aggregate video reward $R(\mathbf{V})$ is the mean of $R_\tau$ across all sampled pairs. 

\subsection{Policy Optimization}
\label{sec:rl-framework}

We optimize the geometry-consistency reward using a customized variant of Flow-GRPO~\cite{liu2025flow}, which extends Group Relative Policy Optimization (GRPO)~\cite{shao2024deepseekmath} to continuous-time flow matching models. We detail our specialized adaptation for high-resolution video generation below.

\paragraph{Flow Matching Background.}
Our video generator builds on the Rectified Flow~\cite{liu2023flow} framework. Given a pristine data sample $\mathbf{x}_0\sim p_{\text{data}}$ and standard Gaussian noise $\boldsymbol{\epsilon}\sim\mathcal{N}(\mathbf{0},\mathbf{I})$, the noised latent state at time $t\in[0,1]$ is defined as:
\begin{equation}
\mathbf{x}_t = (1-t)\,\mathbf{x}_0 + t\,\boldsymbol{\epsilon}.
\label{eq:rectified-flow}
\end{equation}
A neural network $\mathbf{v}_\theta(\mathbf{x}_t,t)$ is trained to predict the velocity field governing this trajectory by minimizing the standard flow matching objective~\cite{lipman2023flow}:
\begin{equation}
\mathcal{L}_{\text{FM}}(\theta) = \mathbb{E}_{t,\,\mathbf{x}_0,\,\boldsymbol{\epsilon}}\!\left[\big\|\mathbf{v}_\theta(\mathbf{x}_t,t) - (\boldsymbol{\epsilon} - \mathbf{x}_0)\big\|^2\right].
\label{eq:flow-matching}
\end{equation}

\paragraph{Denoising as an MDP.}
Following recent RL fine-tuning paradigms~\cite{black2023training,liu2025flow}, we formulate the iterative denoising process as a Markov Decision Process (MDP). The state at step $t$ (we call the discretized steps here $t$ as well to enhance readability) is $s_t = (\mathbf{c}, t, \mathbf{x}_t)$, where $\mathbf{c}$ denotes the conditioning signals (e.g., text prompt and camera parameters). The action is the subsequent denoised state $a_t = \mathbf{x}_{t-1}$, governed by the policy $\pi_\theta(a_t\mid s_t) = p_\theta(\mathbf{x}_{t-1}\mid\mathbf{x}_t,\mathbf{c})$. Crucially, the environment only issues a reward at the terminal step: $R(s_0, a_0) = r(\mathbf{x}_0, \mathbf{c})$, corresponding to our geometry-consistency reward detailed in \S\ref{sec:geometry-reward}.

\paragraph{GRPO Objective and Noise Synchronization.}
Given a prompt $\mathbf{c}$, GRPO samples a group of $G$ candidate videos $\{\mathbf{x}_0^i\}_{i=1}^G$ alongside their respective denoising trajectories. In standard diffusion generation, initial noise is sampled independently for each candidate. However, we empirically find that initializing all $G$ candidates with the \textit{exact same latent noise} $\boldsymbol{\epsilon}$ is critical for stable policy optimization. By locking the initial noise, the behavioral divergence within the group stems strictly from the stochasticity of the policy rather than random initial latent structures. Despite the identical initial noise, the stochasticity injected during the reverse SDE sampling process (see Eq.~\ref{eq:sde-update}) ensures that the $G$ trajectories diverge into distinct candidate videos, allowing us to compute meaningful group-wise advantages. This drastically reduces the variance of the group-normalized advantage:
\begin{equation}
\hat{A}_i = \frac{r(\mathbf{x}_0^i,\mathbf{c}) - \mu_G}{\sigma_G},
\label{eq:advantage}
\end{equation}
where $\mu_G$ and $\sigma_G$ are the mean and standard deviation of the terminal rewards within the $G$-sized group. The policy is then updated by maximizing a clipped surrogate objective, regularized by a KL divergence penalty against the reference (pre-trained) policy $\pi_{\text{ref}}$:
\begin{equation}
\mathcal{L}_{\text{GRPO}}(\theta) = -\,\mathbb{E}_{i,t}\!\left[\min\!\left(
r_t^i(\theta)\,\hat{A}_i,\;
\operatorname{clip}\bigl(r_t^i(\theta),\,1{-}\epsilon_{\mathrm{clip}},\,1{+}\epsilon_{\mathrm{clip}}\bigr)\hat{A}_i
\right)\right],
\label{eq:grpo-loss}
\end{equation}
where $r_t^i(\theta) = \pi_\theta(\mathbf{x}_{t-1}^i\mid\mathbf{x}_t^i,\mathbf{c})\,/\,\pi_{\theta_{\text{old}}}(\mathbf{x}_{t-1}^i\mid\mathbf{x}_t^i,\mathbf{c})$ represents the per-step importance sampling ratio, and $\epsilon_{\mathrm{clip}}$ bounds the policy update.

\paragraph{ODE-to-SDE Conversion.}
Because standard flow matching utilizes a deterministic ODE solver, calculating the per-step log-probabilities $\log\pi_\theta$ required by Eq.~(\ref{eq:grpo-loss}) is intractable, and exploration is inherently stifled. We bypass this by converting the deterministic ODE into an equivalent Stochastic Differential Equation (SDE) that strictly preserves the marginal distributions at all timesteps~\cite{liu2025flow,song2020score}. Discretizing via the Euler-Maruyama method yields the stochastic update:
\begin{equation}
\mathbf{x}_{t-\Delta t} = \mathbf{x}_t - \left[\mathbf{v}_\theta(\mathbf{x}_t,t) + \frac{\sigma_t^2}{2t}\bigl(\mathbf{x}_t + (1{-}t)\,\mathbf{v}_\theta(\mathbf{x}_t,t)\bigr)\right]\Delta t + \sigma_t\sqrt{\Delta t}\;\mathbf{z},
\label{eq:sde-update}
\end{equation}
where $\mathbf{z}\sim\mathcal{N}(\mathbf{0},\mathbf{I})$ and $\sigma_t = a\sqrt{t/(1-t)}$ parameterizes the injected noise level. This transformation re-casts the transition probability $\pi_\theta(\mathbf{x}_{t-1}\mid\mathbf{x}_t,\mathbf{c})$ as an isotropic Gaussian, yielding closed-form, differentiable expressions for both $r_t^i(\theta)$ and the KL divergence.

\paragraph{Full-resolution Truncated Backpropagation.}
Prior implementations of RL for video generation frequently execute training on heavily downsampled spatial resolutions (e.g., operating on $H/2 \times W/2$ grids) to mitigate the extreme VRAM demands of backpropagating through time. In contrast, our geometry and semantic reward pipeline relies heavily on the high-frequency visual details necessary for precise depth and optical flow estimation. Thus, we enforce optimization directly at \textit{full spatial resolution}, maintaining the full latent tensor dimensions $\mathbf{x}_t \in \mathbb{R}^{C \times F \times H \times W}$ throughout the policy update.

To render this full-resolution RL computationally tractable, we introduce a truncated backpropagation strategy. Let $\mathcal{T} = \{t_K, t_{K-1}, \dots, t_1\}$ denote the discrete sequence of reverse-time integration steps, spanning from pure noise at $t_K=1.0$ to the final generated sample at $t_1=0$. During the online rollout phase, we generate the $G$ candidate videos using the full sequence of $K=40$ SDE denoising steps, guaranteeing the high visual fidelity required by our reward. 
 
While the details are useful for precise depth and optical flow, empirical analysis reveals that the global 3D geometry, camera trajectory, and foundational structural layout of the scene are overwhelmingly established during the initial phase of the reverse trajectory~\cite{bahmani2025ac3d}. Exploiting this bias, we partition the sampling schedule into two disjoint subsets: a gradient-tracked early phase $\mathcal{T}_{\text{grad}} = \{t_K, \dots, t_{K-M+1}\}$ and an inference-only late phase $\mathcal{T}_{\text{no-grad}} = \{t_{K-M}, \dots, t_1\}$, where we set $M=20$. We then restrict the policy objective to evaluate expectations exclusively over the early generative phase:
\begin{equation}
\tilde{\mathcal{L}}_{\text{GRPO}}(\theta) = -\,\mathbb{E}_{i} \!\left[ \sum_{t \in \mathcal{T}_{\text{grad}}} \min\!\left( r_t^i(\theta)\,\hat{A}_i,\; \operatorname{clip}\bigl(r_t^i(\theta),\,1{-}\epsilon_{\mathrm{clip}},\,1{+}\epsilon_{\mathrm{clip}}\bigr)\hat{A}_i \right) \right].
\label{eq:grpo-truncated}
\end{equation}
In practice, the backpropagation step takes the majority of the time, so sampling 40 steps while computing gradients for only 20 steps adds little overhead.

\begin{table*}[t]
\centering
\caption{\textbf{Quantitative comparison of geometric consistency}. We compare against Wan2.1, VideoRepa, and additional baseline methods across a $2 \times 2$ matrix of motion regimes: Static vs. Dynamic scenes, and Simple vs. Complex description. Arrows indicate whether higher ($\uparrow$) or lower ($\downarrow$) is better.}
\label{tab:main_results}
\begin{tabular}{ll ccc c ccc}
\toprule
& & \multicolumn{3}{c}{\textbf{Simple}} & & \multicolumn{3}{c}{\textbf{Complex}} \\
\cmidrule{3-5} \cmidrule{7-9}
\textbf{Type} & \textbf{Method} & MEt3R~$\downarrow$ & Sampson~$\downarrow$ & Gemini~$\uparrow$ & & MEt3R~$\downarrow$ & Sampson~$\downarrow$ & Gemini~$\uparrow$ \\
\midrule
\multirow{6}{*}{\textbf{Static}} 
& CogVideoX-1.5    & 5.768 & 1.415 & 8.431 & & 6.412 & 3.169 &  8.440 \\
& VideoGPA  & 6.110 & 1.477 & 8.411 & & 6.882 & 3.274 & 8.425 \\
& GeoVideo  & 12.542 & 8.853 & 7.142 & & 12.838 & 8.024 & 7.336 \\
& VideoRepa & 4.870 & 1.167 & 8.226 & & 6.441 & 2.562 & 8.220 \\
& Wan2.1    & 5.648 & 2.153 & 8.461 & & 5.247 & 3.377 & 8.434 \\
& SFT       & 6.124 & 2.577 & 8.283 & & 5.914 & 3.788 & 8.212 \\
& Ours      & \textbf{3.209} & \textbf{1.096} & \textbf{8.466} & & \textbf{2.498} & \textbf{1.488} & \textbf{8.608} \\
\midrule
\multirow{6}{*}{\textbf{Dynamic}} 
& CogVideoX-1.5    & 19.635 & 7.336 & 8.318 & & 9.021 & \textbf{5.020} & 8.524 \\
& VideoGPA  & 18.840 & 7.259 & 8.342 & & 9.872 & 5.183 & 8.551 \\
& GeoVideo  & 13.071 & 9.073 & 7.235 & & 14.954 & 9.134 & 7.498 \\
& VideoRepa & 10.058 & \textbf{3.586} & 8.200 & & 9.843 & 7.456 & 8.392 \\
& Wan2.1    & 12.292 & 7.351 & 8.279 & & 11.029 & 10.577 & 8.476 \\
& SFT       & 14.407 & 7.838 & 8.118 & & 12.743 & 11.291 & 8.367 \\
& Ours      & \textbf{5.638} & 3.980 & \textbf{8.631} & & \textbf{8.030} & 5.654 & \textbf{8.658} \\
\bottomrule
\end{tabular}
\end{table*}
\begin{figure*}[p]
    \centering
    \includegraphics[width=1.0\linewidth]{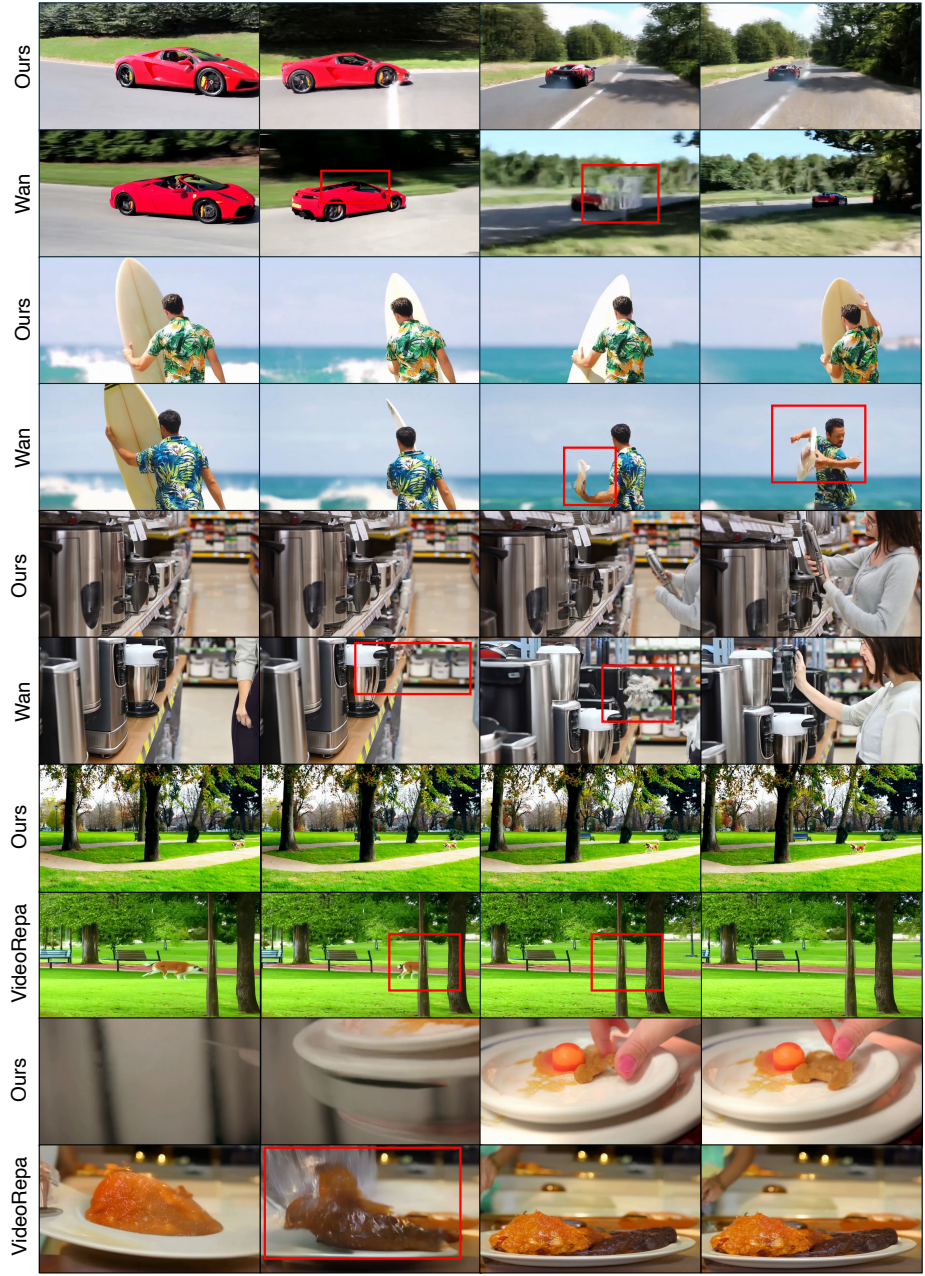}
    \caption{\textbf{Qualitative comparison.} Each row-pair shows sampled frames from a generated video for a different prompt, comparing a baseline model (top) with our fine-tuned model (bottom). The baselines exhibit a range of inconsistencies: objects dissolve or vanish (examples 1, 4), scene layouts shift unexpectedly (ex.~3), identities drift over time (ex.~5), and object structures morph beyond recognition (ex.~2, 3). Our method maintains consistent object permanence, structural integrity, and color stability across all frames. For a better comparison, please view the videos in the supplement.}
    \label{fig:qualitative}
\end{figure*}
\section{Experiments}
We evaluate whether the proposed geometry-consistency reward, combined with Flow-GRPO fine-tuning, improves the spatial coherence of pretrained video diffusion models. We assess performance on both static and dynamic video settings, measure quality preservation, and ablate key reward design choices. 

\paragraph{Experimental Setup.}
We apply our geometry-consistency reward to fine-tune Wan2.1-1.3B~\cite{wan2025wan} using Flow-GRPO. For our training, we use 16 NVIDIA H100 GPUs for sampling and optimization. During the online rollout phase, each GPU processes 4 prompts, generating a group of $G=4$ candidate videos per prompt. During the optimization phase, we use a per-GPU micro-batch size of 2, resulting in an effective global batch size of 128. We train our model for 300 steps, and optimize LoRA~\cite{hu2022lora} weights using AdamW~\cite{kingma2014adam}. The training takes around 24 hours. For each run, the evaluated checkpoint is chosen as the one with the best validation set scores. For evaluation, we follow VBench~\cite{huang2024vbench} and sample each prompt with 5 different initial noises to reduce the influence of the initial latent structure.

\paragraph{Baselines.}
We compare against the unmodified base model Wan2.1-1.3B and an aligned baseline, VideoRepa~\cite{zhang2025videorepa}. While VideoRepa is built on CogVideo-X-2B~\cite{yang2024cogvideox}, it represents the strongest available representation-alignment baseline for physical plausibility in this class of models. Its base architecture features a comparable parameter capacity to ours, making it the most appropriate system-level comparison to evaluate the effectiveness of our optimization. We additionally evaluate recently released geometry-aware baselines, VideoGPA~\cite{du2026videogpadistillinggeometrypriors} and GeoVideo~\cite{bai2025geovideo}, using their official code and checkpoints---both are based on CogVideoX-1.5-5B. Finally, we train a supervised fine-tuning (SFT) control on the same $5{,}000$ prompts used by GeoFlow and their corresponding videos.

\paragraph{Datasets.}
We evaluate on a curated set of 110 unique prompts, yielding 550 evaluated video samples across all runs. This scale is roughly comparable to the consistency evaluation split introduced in VBench2~\cite{zheng2025vbench2}. To construct this set, we sourced prompts from OpenVid-1M~\cite{nan2024openvid} and DL3DV~\cite{ling2024dl3dv}, augmented with rewrites generated by Gemini 3.0-Flash. We then partitioned these prompts into four roughly equal-sized categories based on expected content dynamics (static vs.\ dynamic) and prompt length (simple vs.\ complex). This allows a granular analysis of model performance across varying motion regimes and conditioning complexity. Examples of prompts are provided in the supplementary material.

\paragraph{Metrics.}
We measure consistency using established metrics for static videos, specifically MEt3R~\cite{asim2025met3r} and the Sampson Error~\cite{sampson1982fitting}. While MEt3R and the Sampson Error are traditionally formulated for static scenes, we adapt them for dynamic video evaluation through tight temporal subsampling. By evaluating frame pairs at narrow intervals (a stride of 4 frames), we constrain the magnitude of independent foreground motion. This approximates a quasi-static scene between evaluated frames, making the evaluation more accurate. We complement these with a holistic consistency score generated by Gemini 3.0-Flash (prompt details in supplementary material). For visual quality preservation, we report standard VBench~\cite{huang2024vbench} attributes including background consistency, aesthetic quality, temporal flickering, and motion smoothness.

\subsection{Geometric Consistency}
As shown in Table~\ref{tab:main_results},
on static scenes with simple prompts, our method reduces MEt3R from 5.648 to 3.209 (a 43\% reduction) and Sampson distance from 2.153 to 1.096 (49\% reduction), 
indicating substantially improved cross-view geometric agreement. Gains are even more pronounced on complex static prompts, where MEt3R drops from 5.247 to 2.498 (52\% reduction). Compared to VideoRepa, which targets physical plausibility, our method achieves consistently lower MEt3R across all settings (e.g., 3.209 vs.\ 4.870 on simple static scenes) while also obtaining higher Gemini consistency scores (8.466 vs.\ 8.226). On dynamic scenes, we observe a similar trend. Our method roughly halves the Sampson score for both simple and complex prompts and significantly improves MEt3R over Wan and VideoRepa. It also achieves the highest Gemini score in all rubrics. Figure~\ref{fig:qualitative} provides a qualitative comparison: both baselines exhibit noticeable drift and artifacts, while our method produces outputs with consistent scene structure across the full camera trajectory. Additionally, the qualitative results show that the overall aesthetic, camera movement, and object movement remain similar, indicating that our optimization improves inconsistent parts of the scene without shifting toward a completely different scene.

Table~\ref{tab:backbone_generalization} further tests whether these gains depend on the Wan2.1 backbone, while Table~\ref{tab:main_results} compares against the additional geometry-aware baselines and the SFT control. Since CogVideoX-2B is not a flow-matching model, we apply the same reward using standard DDPO~\cite{black2023training} for video on top of a DDIM sampler. This reduces MEt3R from 9.58 to 6.01 and Sampson error from 4.74 to 3.61, while slightly improving aesthetic quality from 0.571 to 0.580, showing that the reward transfers to a second video model.

The additional baselines highlight that the gains are not explained by generic supervised tuning or by existing geometry-aware fine-tuning alone. SFT on the same $5{,}000$ training prompts worsens the base model, increasing average MEt3R by 0.22 and Sampson error by 0.13. This suggests that, at this data scale, supervised fine-tuning overfits the generator to the training prompt distribution instead of improving geometric consistency more generally. VideoGPA changes the outputs only mildly relative to its CogVideoX-1.5 base model, which is reflected in its small average metric changes (8.85 to 8.41 MEt3R, 4.17 to 4.11 Sampson). GeoVideo performs substantially worse on our benchmark, especially in dynamic scenes (13.70 MEt3R, 9.45 Sampson), and its generations often contain strong visual artifacts. One likely reason is that fine-tuning on the GeoVideo prompt and caption distribution does not transfer well to our prompts, which are sourced from different datasets and captioned with a different style.

\begin{table}[t]
    \centering
    \caption{\textbf{Generalization to another video backbone.} GeoFlow improves geometric consistency when applying the same reward to CogVideoX-2B with DDPO on top of a DDIM sampler.}
    \label{tab:backbone_generalization}
    \small
    \setlength{\tabcolsep}{5pt}
    \begin{tabular}{lccc}
        \toprule
        \textbf{Method} & \textbf{MEt3R} $\downarrow$ & \textbf{Sampson} $\downarrow$ & \textbf{Aes.} $\uparrow$ \\
        \midrule
        CogVideoX-2B & 9.58 & 4.74 & 0.571 \\
        CogVideoX-2B + GeoFlow & \textbf{6.01} & \textbf{3.61} & \textbf{0.580} \\
        \bottomrule
    \end{tabular}
\end{table}

\subsection{Quality Preservation}

A key concern with any alignment procedure is whether optimizing for a specific reward degrades other aspects of generation quality. Table~\ref{tab:vbench_1} reports standard VBench attributes measuring perceptual quality, including the additional baselines.

Our method preserves or improves quality across all measured attributes relative to the unaligned Wan2.1: aesthetic quality slightly increases (0.589 vs.\ 0.582), temporal flickering remains near-perfect (0.996 vs.\ 0.994), and background consistency improves from 0.958 to 0.971. Motion smoothness remains comparable (0.987 vs.\ 0.971). These results demonstrate that the geometry-consistency reward does not compromise visual fidelity.

In contrast, VideoRepa exhibits degraded aesthetic quality (0.477 vs.\ 0.608), lower background consistency (0.945 vs.\ 0.966), and increased temporal flickering (0.983 vs.\ 0.989). While some of this disparity may stem from noise in the evaluation protocol, the severity of some artifacts suggests that its representation-alignment objective may over-constrain the generator at the expense of visual quality. Our method achieves a more favorable consistency--quality trade-off.

\begin{table}[t]
    \centering
    \caption{\textbf{VBench 1.0 evaluation.} Our method preserves perceptual quality across all standard VBench attributes while achieving the consistency gains reported in Table~\ref{tab:main_results}.}
    \label{tab:vbench_1}
    \small
    \setlength{\tabcolsep}{4pt}
    \begin{tabular}{lcccc}
        \toprule
        \textbf{Method} & \textbf{BG~Consist.$~\uparrow$} & \textbf{Aes.~Qual.$~\uparrow$} & \textbf{Temp.~Flick.$~\uparrow$} & \textbf{Mot.~Smooth.$~\uparrow$} \\
        \midrule
        CogVideoX-2B & 0.966 & 0.571 & 0.989 & 0.977 \\
        Wan2.1 & 0.958 & 0.582 & 0.994 & 0.971 \\
        VideoRepa & 0.945 & 0.477 & 0.983 & \textbf{0.990} \\
        VideoGPA & 0.961 & \textbf{0.600} & 0.966 & 0.974 \\
        GeoVideo & \textbf{0.978} & 0.501 & 0.949 & 0.981 \\
        SFT & 0.942 & 0.566 & 0.986 & 0.964 \\
        Ours & \underline{0.971} & \underline{0.589} & \textbf{0.996} & \underline{0.987} \\
        \bottomrule
    \end{tabular}
\end{table}

\subsection{Ablations}
In this section, we ablate the main design choices of our method, including the reward function and the adaptations made to Flow-GRPO.

\paragraph{Reward Function.}
We ablate the reward design to isolate which geometric signals most effectively drive consistency improvements. Table~\ref{tab:ablation_rewards} compares different reward formulations: our combined reward, the MEt3R component alone, the Epipolar reward (this reward was incapable of producing stable rewards under the ablation settings so we trained it with distinct initial noises), and a Gemini-based consistency signal.

The Epipolar reward achieves a low Sampson distance (3.091) by construction, as it directly optimizes the epipolar constraint. The Gemini reward, while improving Gemini scores (8.501 vs.\ 8.426), provides only modest geometric improvements (MEt3R of 7.260 vs.\ 8.212), suggesting that holistic assessment signals lack the geometric specificity needed for strong cross-view coherence. MEt3R alone achieves comparable MEt3R (5.716 vs.\ 5.730) and Sampson distance (3.256 vs.\ 3.091) to the Epipolar reward while yielding the highest aesthetic quality (0.594). Our combined reward achieves the best overall consistency with the lowest MEt3R (4.594) and Sampson distance (2.709) while maintaining strong aesthetic quality (0.589), demonstrating that combining complementary geometric signals produces the most effective reward. In absolute MEt3R reduction over the base model, the full reward improves 44\% more than directly optimizing MEt3R alone, confirming that the gain comes from the decoupled reward rather than simply using a stronger geometry estimator.

\begin{table}[t]
    \centering
    \caption{\textbf{Reward function ablation.} Comparison of different reward functions for RL fine-tuning. The best result is shown in bold.}
    \label{tab:ablation_rewards}
    \small
    \setlength{\tabcolsep}{4pt}
    \begin{tabular}{lcccc}
        \toprule
        \textbf{Reward Type} & \textbf{MEt3R} $\downarrow$ & \textbf{Sampson} $\downarrow$ & \textbf{Gemini} $\uparrow$  & \textbf{Aes.~Qual.} $\uparrow$ \\
        \midrule
        Base Model & 8.212 & 5.453 & 8.426 & 0.582 \\
        Gemini & 7.260 & 4.412 & 8.501 & 0.584 \\
        $\text{Epipolar}^*$ & 5.730 & 3.091 & 8.557 & 0.589 \\
        MEt3R & 5.716 & 3.256 & 8.527 & \textbf{0.594} \\
        Ours & \textbf{4.594} & \textbf{2.709} & \textbf{8.580} & 0.589\\
        \bottomrule
    \end{tabular}
\end{table}

\paragraph{Reinforcement Learning Adaptations.}
We ablate the key optimization components of our Flow-GRPO adaptations (Table~\ref{tab:ablation_rl}). All variants use our combined reward, isolating the effect of each change.

Removing full-resolution reward computation (w/o Full Res) degrades MEt3R from 4.594 to 5.378 and Sampson distance from 2.709 to 3.296, confirming that evaluating geometric consistency at full resolution is important for providing informative gradients. Removing window-based optimization (w/o Truncated Optimization)---where only 20 steps are denoised and fully used for optimization---yields a similar degradation (MEt3R of 5.401, Sampson of 3.426), indicating that this approach is critical for stable training. Training with distinct initial noises also leads to similar degradations in all dimensions. The full configuration combines both techniques and achieves the best results across all metrics, including the highest Gemini score (8.580) and aesthetic quality (0.589).

\begin{table}[t]
    \centering
    \caption{\textbf{Optimization ablation.} Impact of removing components of our method during training. The best result is shown in bold.}
    \label{tab:ablation_rl}
    \small
    \setlength{\tabcolsep}{1pt}
    \begin{tabular}{lcccc}
        \toprule
        \textbf{Configuration} & \textbf{MEt3R} $\downarrow$ & \textbf{Sampson} $\downarrow$ & \textbf{Gemini} $\uparrow$ & \textbf{Aes.~Qual.} $\uparrow$ \\
        \midrule
        Base Model               & 8.212 & 5.453 & 8.426 & 0.582 \\
        w/o Full Resolution  & 5.378  & 3.296  & 8.511  &  0.584 \\
        w/o Truncated Optimization & 5.401  & 3.426  & 8.467 & 0.587  \\
        w/o Same Latents & 5.407 & 3.425 & 8.447 & 0.582 \\
        Ours & \textbf{4.594} & \textbf{2.709}  & \textbf{8.580}  & \textbf{0.589} \\
        \bottomrule
    \end{tabular}
\end{table}
\section{Conclusion}
\label{sec:conclusion}
In this work, we introduced a framework to improve the temporal and geometric consistency of pretrained text-to-video diffusion models. By formulating a dense, self-supervised geometry-consistency reward, we integrate the evaluation of rigid, camera-induced background motion with identity preservation for dynamic foreground objects within a joint continuous objective. Paired with our tailored adaptations to the Flow-GRPO reinforcement learning algorithm---including synchronized noise initialization and full-resolution truncated backpropagation---our method provides a robust alignment signal without requiring costly 3D or 4D ground-truth data. Our evaluations demonstrate that this approach reduces common generative artifacts, such as background wobble, structural deformation, and texture drift, while maintaining the perceptual quality and diversity of the underlying base model.

\paragraph{Limitations.}
Despite its effectiveness, our approach presents a few notable limitations. First, the reliability of our reward signal inherently bounds the policy optimization; because we rely on off-the-shelf priors (e.g., Depth Anything 3 and WAFT), scenarios that induce severe failure modes in these models can inject noise into the reward. Second, while our truncated backpropagation strategy mitigates peak memory usage, full-resolution reinforcement learning through time remains highly computationally intensive compared to standard supervised fine-tuning or low-resolution RL. Finally, our bipartite motion assumption (a rigid background with independent dynamic objects) may inappropriately penalize the generation of globally non-rigid or morphing environments, such as turbulent water, large-scale fire, or abstract artistic scenes.

\paragraph{Future Work.}
There are several exciting avenues for future research. One immediate direction is extending this framework to autoregressive or ultra-long-form video generation---where compounding geometric drift and identity loss are the primary failure modes. Additionally, investigating more parameter-efficient RL formulations or distilling the optimization process into a lightweight, geometry-aware discriminator could substantially reduce computational overhead and democratize the training of geometrically coherent video foundation models.

\bibliographystyle{splncs04}
\bibliography{main}

\begin{thebibliography}{10}
\providecommand{\url}[1]{\texttt{#1}}
\providecommand{\urlprefix}{URL }
\providecommand{\doi}[1]{https://doi.org/#1}

\bibitem{asim2025met3r}
Asim, M., Wewer, C., Wimmer, T., Schiele, B., Lenssen, J.E.: Met3r: Measuring
  multi-view consistency in generated images. In: Proceedings of the IEEE/CVF
  Conference on Computer Vision and Pattern Recognition. pp. 6034--6044 (2025)

\bibitem{bahmani2025ac3d}
Bahmani, S., Skorokhodov, I., Qian, G., Siarohin, A., Menapace, W.,
  Tagliasacchi, A., Lindell, D.B., Tulyakov, S.: {AC3D: Analyzing and Improving
  3D Camera Control in Video Diffusion Transformers}. In: Proceedings of the
  IEEE/CVF Conference on Computer Vision and Pattern Recognition (CVPR). pp.
  22875--22889 (2025)

\bibitem{bahmani2024vd3d}
Bahmani, S., Skorokhodov, I., Siarohin, A., Menapace, W., Qian, G.,
  Vasilkovsky, M., Lee, H.Y., Wang, C., Zou, J., Tagliasacchi, A., et~al.:
  {VD3D: Taming Large Video Diffusion Transformers for 3D Camera Control}.
  arXiv preprint arXiv:2407.12781  (2024)

\bibitem{bai2024syncammaster}
Bai, J., Xia, M., Wang, X., Yuan, Z., Fu, X., Liu, Z., Hu, H., Wan, P., Zhang,
  D.: {SyncamMaster: Synchronizing Multi-Camera Video Generation from Diverse
  Viewpoints}. arXiv preprint arXiv:2412.07760  (2024)

\bibitem{bai2025geovideo}
Bai, Y., Fang, S., Yu, C., Wang, F., Huang, Q.: Geovideo: Introducing geometric
  regularization into video generation model. arXiv preprint arXiv:2512.03453
  (2025)

\bibitem{batifol2025flux}
Batifol, S., Blattmann, A., Boesel, F., Consul, S., Diagne, C., Dockhorn, T.,
  English, J., English, Z., Esser, P., Kulal, S., et~al.: Flux. 1 kontext: Flow
  matching for in-context image generation and editing in latent space. arXiv
  e-prints pp. arXiv--2506 (2025)

\bibitem{bengtson2025geometric}
Bengtson, J., Nilsson, D., Kahl, F.: Geometric consistency refinement for
  single image novel view synthesis via test-time adaptation of diffusion
  models. In: Proceedings of the Computer Vision and Pattern Recognition
  Conference. pp. 6399--6408 (2025)

\bibitem{black2023training}
Black, K., Janner, M., Du, Y., Kostrikov, I., Levine, S.: {Training Diffusion
  Models with Reinforcement Learning}. arXiv preprint arXiv:2305.13301  (2023)

\bibitem{videoworldsimulators2024}
Brooks, T., Peebles, B., Holmes, C., DePue, W., Guo, Y., Jing, L., Schnurr, D.,
  Taylor, J., Luhman, T., Luhman, E., Ng, C., Wang, R., Ramesh, A.: Video
  generation models as world simulators  (2024),
  \url{https://openai.com/research/video-generation-models-as-world-simulators}

\bibitem{bruce2024genie}
Bruce, J., Dennis, M.D., Edwards, A., Parker-Holder, J., Shi, Y., Hughes, E.,
  Lai, M., Mavalankar, A., Steigerwald, R., Apps, C., et~al.: {Genie:
  Generative Interactive Environments}. In: Proceedings of the International
  Conference on Machine Learning (ICML) (2024)

\bibitem{cai2025mixture}
Cai, S., Yang, C., Zhang, L., Guo, Y., Xiao, J., Yang, Z., Xu, Y., Yang, Z.,
  Yuille, A., Guibas, L., et~al.: Mixture of contexts for long video
  generation. arXiv preprint arXiv:2508.21058  (2025)

\bibitem{cao2025mvgenmaster}
Cao, C., et~al.: {MVGenMaster: Scaling Multi-View Generation from Any Image via
  3D Priors}. In: Proceedings of the IEEE/CVF Conference on Computer Vision and
  Pattern Recognition (CVPR) (2025),
  \url{https://openaccess.thecvf.com/content/CVPR2025/papers/Cao_MVGenMaster_Scaling_Multi-View_Generation_from_Any_Image_via_3D_Priors_CVPR_2025_paper.pdf}

\bibitem{chan2022efficient}
Chan, E.R., Lin, C.Z., Chan, M.A., Nagano, K., Pan, B., De~Mello, S., Gallo,
  O., Guibas, L.J., Tremblay, J., Khamis, S., et~al.: {Efficient Geometry-Aware
  3D Generative Adversarial Networks}. In: Proceedings of the IEEE/CVF
  Conference on Computer Vision and Pattern Recognition (CVPR). pp.
  16123--16133 (2022)

\bibitem{chan2021pi}
Chan, E.R., Monteiro, M., Kellnhofer, P., Wu, J., Wetzstein, G.: {pi-GAN:
  Periodic Implicit Generative Adversarial Networks for 3D-Aware Image
  Synthesis}. In: Proceedings of the IEEE/CVF Conference on Computer Vision and
  Pattern Recognition (CVPR). pp. 5799--5809 (2021)

\bibitem{chan2023generative}
Chan, E.R., Nagano, K., Chan, M.A., Bergman, A.W., Park, J.J., Levy, A.,
  Aittala, M., De~Mello, S., Karras, T., Wetzstein, G.: {Generative Novel View
  Synthesis with 3D-Aware Diffusion Models}. In: Proceedings of the IEEE/CVF
  International Conference on Computer Vision (ICCV). pp. 4217--4229 (2023)

\bibitem{chen2024diffusion}
Chen, B., Mart{\'\i}~Mons{\'o}, D., Du, Y., Simchowitz, M., Tedrake, R.,
  Sitzmann, V.: Diffusion forcing: Next-token prediction meets full-sequence
  diffusion. Advances in Neural Information Processing Systems  \textbf{37},
  24081--24125 (2024)

\bibitem{chen2024-3dadapter}
Chen, H., Shen, B., Liu, Y., Shi, R., Zhou, L., Lin, C.Z., Gu, J., Su, H.,
  Wetzstein, G., Guibas, L.: {3D-Adapter: Geometry-Consistent Multi-View
  Diffusion for High-Quality 3D Generation}. arXiv preprint arXiv:2410.18974
  (2024), \url{https://arxiv.org/abs/2410.18974}

\bibitem{chen2024meshxl}
Chen, S., Chen, X., Pang, A., Zeng, X., Cheng, W., Fu, Y., Yin, F., Wang, B.,
  Yu, J., Yu, G., et~al.: {MeshXL: Neural Coordinate Field for Generative 3D
  Foundation Models}. In: Advances in Neural Information Processing Systems
  (NeurIPS). vol.~37, pp. 97141--97166 (2024)

\bibitem{chen2024mvsplat}
Chen, Y., Xu, H., Zheng, C., Zhuang, B., Pollefeys, M., Geiger, A., Cham, T.J.,
  Cai, J.: {MVSplat: Efficient 3D Gaussian Splatting from Sparse Multi-View
  Images}. In: Proceedings of the European Conference on Computer Vision
  (ECCV). pp. 370--386. Springer (2024)

\bibitem{chen20253dtopia}
Chen, Z., Tang, J., Dong, Y., Cao, Z., Hong, F., Lan, Y., Wang, T., Xie, H.,
  Wu, T., Saito, S., et~al.: {3DTopia-XL: Scaling High-Quality 3D Asset
  Generation via Primitive Diffusion}. In: Proceedings of the IEEE/CVF
  Conference on Computer Vision and Pattern Recognition (CVPR). pp.
  26576--26586 (2025)

\bibitem{chou2025generating}
Chou, G., Zhang, K., Bi, S., Tan, H., Xu, Z., Luan, F., Hariharan, B., Snavely,
  N.: Generating 3d-consistent videos from unposed internet photos. In:
  Proceedings of the Computer Vision and Pattern Recognition Conference. pp.
  27934--27945 (2025)

\bibitem{du2026videogpadistillinggeometrypriors}
Du, H., Ye, J., Cong, X., Li, R., Ni, J., Agarwal, A., Zhou, Z., Li, Z.,
  Balestriero, R., Wang, Y.: Videogpa: Distilling geometry priors for
  3d-consistent video generation (2026), \url{https://arxiv.org/abs/2601.23286}

\bibitem{edelstein2025sharpit}
Edelstein, Y., Patashnik, O., Cohen-Bar, D., Zelnik-Manor, L.: {Sharp-It: A
  Multi-view to Multi-view Diffusion Model for 3D Synthesis and Manipulation}.
  In: Proceedings of the IEEE/CVF Conference on Computer Vision and Pattern
  Recognition (CVPR) (2025),
  \url{https://openaccess.thecvf.com/content/CVPR2025/papers/Edelstein_Sharp-It_A_Multi-view_to_Multi-view_Diffusion_Model_for_3D_Synthesis_CVPR_2025_paper.pdf}

\bibitem{gao2022get3d}
Gao, J., Shen, T., Wang, Z., Chen, W., Yin, K., Li, D., Litany, O., Gojcic, Z.,
  Fidler, S.: {GET3D: A Generative Model of High Quality 3D Textured Shapes
  Learned from Images}. In: Advances in Neural Information Processing Systems
  (NeurIPS). vol.~35, pp. 31841--31854 (2022)

\bibitem{gao2024cat3d}
Gao, R., Holynski, A., Henzler, P., Brussee, A., Martin-Brualla, R.,
  Srinivasan, P., Barron, J.T., Poole, B.: {Cat3D: Create Anything in 3D with
  Multi-View Diffusion Models}. arXiv preprint arXiv:2405.10314  (2024)

\bibitem{gu2025geco}
Gu, L., Hur, J., Herrmann, C., Zhan, F., Zickler, T., Sun, D., Pfister, H.:
  Geco: A differentiable geometric consistency metric for video generation.
  arXiv preprint arXiv:2512.22274  (2025)

\bibitem{he2024cameractrl}
He, H., et~al.: {CameraCtrl: Enabling Camera Control for Text-to-Video
  Diffusion Models}. arXiv preprint arXiv:2404.02101  (2024),
  \url{https://arxiv.org/abs/2404.02101}

\bibitem{hong2022cogvideo}
Hong, W., Ding, M., Zheng, W., Liu, X., Tang, J.: Cogvideo: Large-scale
  pretraining for text-to-video generation via transformers. arXiv preprint
  arXiv:2205.15868  (2022)

\bibitem{hong2023lrm}
Hong, Y., Zhang, K., Gu, J., Bi, S., Zhou, Y., Liu, D., Liu, F., Sunkavalli,
  K., Bui, T., Tan, H.: {LRM: Large Reconstruction Model for Single Image to
  3D}. arXiv preprint arXiv:2311.04400  (2023),
  \url{https://arxiv.org/abs/2311.04400}

\bibitem{hu2022lora}
Hu, E.J., Shen, Y., Wallis, P., Allen-Zhu, Z., Li, Y., Wang, S., Wang, L.,
  Chen, W., et~al.: Lora: Low-rank adaptation of large language models. Iclr
  \textbf{1}(2), ~3 (2022)

\bibitem{huang2024vbench}
Huang, Z., He, Y., Yu, J., Zhang, F., Si, C., Jiang, Y., Zhang, Y., Wu, T.,
  Jin, Q., Chanpaisit, N., et~al.: Vbench: Comprehensive benchmark suite for
  video generative models. In: Proceedings of the IEEE/CVF Conference on
  Computer Vision and Pattern Recognition. pp. 21807--21818 (2024)

\bibitem{kingma2014adam}
Kingma, D.P., Ba, J.: Adam: A method for stochastic optimization. arXiv
  preprint arXiv:1412.6980  (2014)

\bibitem{kong2024hunyuanvideo}
Kong, W., Tian, Q., Zhang, Z., Min, R., Dai, Z., Zhou, J., Xiong, J., Li, X.,
  Wu, B., Zhang, J., et~al.: Hunyuanvideo: A systematic framework for large
  video generative models. arXiv preprint arXiv:2412.03603  (2024)

\bibitem{kuang2024collaborative}
Kuang, Z., Cai, S., He, H., Xu, Y., Li, H., Guibas, L.J., Wetzstein, G.:
  {Collaborative Video Diffusion: Consistent Multi-Video Generation with Camera
  Control}. In: Advances in Neural Information Processing Systems (NeurIPS).
  vol.~37, pp. 16240--16271 (2024)

\bibitem{kupyn2025epipolar}
Kupyn, O., Manhardt, F., Tombari, F., Rupprecht, C.: Epipolar geometry improves
  video generation models. arXiv preprint arXiv:2510.21615  (2025)

\bibitem{leroy2024grounding}
Leroy, V., Cabon, Y., Revaud, J.: {Grounding Image Matching in 3D with MAST3R}.
  In: Proceedings of the European Conference on Computer Vision (ECCV). pp.
  71--91. Springer (2024)

\bibitem{li2025mixgrpo}
Li, J., Cui, Y., Huang, T., Ma, Y., Fan, C., Yang, M., Zhong, Z.: Mixgrpo:
  Unlocking flow-based grpo efficiency with mixed ode-sde. arXiv preprint
  arXiv:2507.21802  (2025)

\bibitem{li2024era3d}
Li, P., et~al.: {Era3D: High-Resolution Multiview Diffusion Using Efficient
  Rearrangement Attention}. In: Advances in Neural Information Processing
  Systems (NeurIPS) (2024),
  \url{https://proceedings.neurips.cc/paper_files/paper/2024/file/65a723bf7d8dad838c09178270d30e80-Paper-Conference.pdf}

\bibitem{liang2024richhf}
Liang, Y., et~al.: {Rich Human Feedback for Text-to-Image Generation}. In:
  Proceedings of the IEEE/CVF Conference on Computer Vision and Pattern
  Recognition (CVPR) (2024),
  \url{https://openaccess.thecvf.com/content/CVPR2024/papers/Liang_Rich_Human_Feedback_for_Text-to-Image_Generation_CVPR_2024_paper.pdf}

\bibitem{lin2025depth}
Lin, H., Chen, S., Liew, J., Chen, D.Y., Li, Z., Shi, G., Feng, J., Kang, B.:
  Depth anything 3: Recovering the visual space from any views. arXiv preprint
  arXiv:2511.10647  (2025)

\bibitem{ling2024dl3dv}
Ling, L., Sheng, Y., Tu, Z., Zhao, W., Xin, C., Wan, K., Yu, L., Guo, Q., Yu,
  Z., Lu, Y., et~al.: {DL3DV-10K: A Large-Scale Scene Dataset for Deep
  Learning-based 3D Vision}. In: Proceedings of the IEEE/CVF Conference on
  Computer Vision and Pattern Recognition (CVPR). pp. 22166--22176 (2024)

\bibitem{lipman2023flow}
Lipman, Y., Chen, R.T.Q., Ben-Hamu, H., Nickel, M., Le, M.: {Flow Matching for
  Generative Modeling}. arXiv preprint arXiv:2210.02747  (2023)

\bibitem{lipman2022flow}
Lipman, Y., Chen, R.T., Ben-Hamu, H., Nickel, M., Le, M.: Flow matching for
  generative modeling. arXiv preprint arXiv:2210.02747  (2022)

\bibitem{liu2025flow}
Liu, J., Liu, G., Liang, J., Li, Y., Liu, J., Wang, X., Wan, P., Zhang, D.,
  Ouyang, W.: {Flow-GRPO: Training Flow Matching Models via Online RL}. arXiv
  preprint arXiv:2505.05470  (2025)

\bibitem{liu2023zero123}
Liu, R., Wu, R., Van~Hoorick, B., Tokmakov, P., Zakharov, S., Vondrick, C.:
  {Zero-1-to-3: Zero-shot One Image to 3D Object}. In: Proceedings of the
  IEEE/CVF International Conference on Computer Vision (ICCV). pp. 9298--9309
  (2023), \url{https://arxiv.org/abs/2303.11328}

\bibitem{liu2023flow}
Liu, X., Gong, C., Liu, Q.: {Flow Straight and Fast: Learning to Generate and
  Transfer Data with Rectified Flow}. arXiv preprint arXiv:2209.03003  (2023)

\bibitem{liu2023syncdreamer}
Liu, Y., et~al.: {SyncDreamer: Generating Multiview-Consistent Images from a
  Single-View Image}. arXiv preprint arXiv:2309.03453  (2023),
  \url{https://arxiv.org/abs/2309.03453}

\bibitem{lowe2004distinctive}
Lowe, D.G.: Distinctive image features from scale-invariant keypoints.
  International journal of computer vision  \textbf{60}(2),  91--110 (2004)

\bibitem{nan2024openvid}
Nan, K., Xie, R., Zhou, P., Fan, T., Zheng, Z., Huang, Z., Li, H., Li, J., Li,
  J.: {OpenVid-1M: A Large-Scale High-Quality Dataset for Text-to-Video
  Generation}. arXiv preprint arXiv:2407.02371  (2024)

\bibitem{oquab2023dinov2}
Oquab, M., Darcet, T., Moutakanni, T., Vo, H., Szafraniec, M., Khalidov, V.,
  Fernandez, P., Haziza, D., Massa, F., El-Nouby, A., et~al.: Dinov2: Learning
  robust visual features without supervision. arXiv preprint arXiv:2304.07193
  (2023)

\bibitem{po2025bagger}
Po, R., Chan, E.R., Chen, C., Wetzstein, G.: Bagger: Backwards aggregation for
  mitigating drift in autoregressive video diffusion models. arXiv preprint
  arXiv:2512.12080  (2025)

\bibitem{po2025long}
Po, R., Nitzan, Y., Zhang, R., Chen, B., Dao, T., Shechtman, E., Wetzstein, G.,
  Huang, X.: {Long-Context State-Space Video World Models}. arXiv preprint
  arXiv:2505.20171  (2025)

\bibitem{rafailov2023dpo}
Rafailov, R., Sharma, A., Mitchell, E., Manning, C.D., Ermon, S., Finn, C.:
  Direct preference optimization: Your language model is secretly a reward
  model. In: NeurIPS (2023)

\bibitem{rombach2022high}
Rombach, R., Blattmann, A., Lorenz, D., Esser, P., Ommer, B.: High-resolution
  image synthesis with latent diffusion models. In: Proceedings of the IEEE/CVF
  conference on computer vision and pattern recognition. pp. 10684--10695
  (2022)

\bibitem{sampson1982fitting}
Sampson, P.D.: Fitting conic sections to “very scattered” data: An
  iterative refinement of the bookstein algorithm. Computer graphics and image
  processing  \textbf{18}(1),  97--108 (1982)

\bibitem{seo2023let}
Seo, J., Jang, W., Kwak, M.S., Kim, H., Ko, J., Kim, J., Kim, J.H., Lee, J.,
  Kim, S.: {Let 2D Diffusion Model Know 3D-Consistency for Robust Text-to-3D
  Generation}. arXiv preprint arXiv:2303.07937  (2023)

\bibitem{shao2024grpo}
Shao, Z., Wang, P., Zhu, Q., Xu, R., Song, J., Bi, X., Zhang, H., Zhang, M.,
  Li, Y.K., Wu, Y., Guo, D.: Deepseekmath: Pushing the limits of mathematical
  reasoning in open language models. In: arXiv (2024)

\bibitem{shao2024deepseekmath}
Shao, Z., Wang, P., Zhu, Q., Xu, R., Song, J., Zhang, M., Li, Y., Wu, Y., Guo,
  D.: {DeepSeekMath: Pushing the Limits of Mathematical Reasoning in Open
  Language Models}. arXiv preprint arXiv:2402.03300  (2024)

\bibitem{shi2023mvdream}
Shi, Y., Wang, P., Ye, J., Long, M., Li, K., Yang, X.: {MVDream: Multi-view
  Diffusion for 3D Generation}. arXiv preprint arXiv:2308.16512  (2023)

\bibitem{shi2023zero123pp}
Shi, Y., Wang, P., Ye, J., Mai, L., Li, K., Yang, X.: {Zero123++: Single Image
  to Consistent Multi-view Diffusion Base Model}. arXiv preprint
  arXiv:2310.15110  (2023), \url{https://arxiv.org/abs/2310.15110}

\bibitem{shue20233d}
Shue, J.R., Chan, E.R., Po, R., Ankner, Z., Wu, J., Wetzstein, G.: {3D Neural
  Field Generation using Triplane Diffusion}. In: Proceedings of the IEEE/CVF
  Conference on Computer Vision and Pattern Recognition (CVPR). pp.
  20875--20886 (2023)

\bibitem{song2025history}
Song, K., Chen, B., Simchowitz, M., Du, Y., Tedrake, R., Sitzmann, V.:
  {History-Guided Video Diffusion}. arXiv preprint arXiv:2502.06764  (2025)

\bibitem{song2020score}
Song, Y., Sohl-Dickstein, J., Kingma, D.P., Kumar, A., Ermon, S., Poole, B.:
  Score-based generative modeling through stochastic differential equations.
  arXiv preprint arXiv:2011.13456  (2020)

\bibitem{tang2024lgm}
Tang, J., Chen, Z., Chen, X., Wang, T., Zeng, G., Liu, Z.: {LGM: Large
  Multi-View Gaussian Model for High-Resolution 3D Content Creation}. In:
  Proceedings of the European Conference on Computer Vision (ECCV) (2024),
  \url{https://arxiv.org/abs/2402.05054}

\bibitem{team2023gemini}
Team, G., Anil, R., Borgeaud, S., Alayrac, J.B., Yu, J., Soricut, R.,
  Schalkwyk, J., Dai, A.M., Hauth, A., Millican, K., et~al.: Gemini: a family
  of highly capable multimodal models. arXiv preprint arXiv:2312.11805  (2023)

\bibitem{genmo2024mochi}
Team, G.: Mochi 1. \url{https://github.com/genmoai/models} (2024)

\bibitem{teed2020raft}
Teed, Z., Deng, J.: Raft: Recurrent all-pairs field transforms for optical
  flow. In: European conference on computer vision. pp. 402--419. Springer
  (2020)

\bibitem{tochilkin2024triposr}
Tochilkin, D., Pankratz, D., Liu, Z., Huang, Z., Letts, A., Li, Y., Liang, D.,
  Laforte, C., Jampani, V., Cao, Y.P.: {TripoSR: Fast 3D Object Reconstruction
  from a Single Image}. arXiv preprint arXiv:2403.02151  (2024),
  \url{https://arxiv.org/abs/2403.02151}

\bibitem{van2024generative}
Van~Hoorick, B., Wu, R., Ozguroglu, E., Sargent, K., Liu, R., Tokmakov, P.,
  Dave, A., Zheng, C., Vondrick, C.: {Generative Camera Dolly: Extreme
  Monocular Dynamic Novel View Synthesis}. In: Proceedings of the European
  Conference on Computer Vision (ECCV). pp. 313--331. Springer (2024)

\bibitem{wallace2024diffusiondpo}
Wallace, B., et~al.: {Diffusion Model Alignment Using Direct Preference
  Optimization}. In: Proceedings of the IEEE/CVF Conference on Computer Vision
  and Pattern Recognition (CVPR) (2024),
  \url{https://openaccess.thecvf.com/content/CVPR2024/papers/Wallace_Diffusion_Model_Alignment_Using_Direct_Preference_Optimization_CVPR_2024_paper.pdf}

\bibitem{wan2025wan}
{Wan Team}: {Wan: Open and Advanced Large-Scale Video Generative Models}. arXiv
  preprint arXiv:2503.20314  (2025)

\bibitem{wang2025pi}
Wang, Y., Zhou, J., Zhu, H., Chang, W., Zhou, Y., Li, Z., Chen, J., Pang, J.,
  Shen, C., He, T.: {$\pi^{3}$: Scalable Permutation-Equivariant Visual
  Geometry Learning}. arXiv preprint arXiv:2507.13347  (2025)

\bibitem{wang2025waft}
Wang, Y., Deng, J.: Waft: Warping-alone field transforms for optical flow.
  arXiv preprint arXiv:2506.21526  (2025)

\bibitem{wiedemer2025video}
Wiedemer, T., Li, Y., Vicol, P., Gu, S.S., Matarese, N., Swersky, K., Kim, B.,
  Jaini, P., Geirhos, R.: Video models are zero-shot learners and reasoners.
  arXiv preprint arXiv:2509.20328  (2025)

\bibitem{wu2025ic}
Wu, F., Wei, J., Li, R., Xu, Y., Li, J., Ye, D., Lin, G.: Ic-world: In-context
  generation for shared world modeling. arXiv preprint arXiv:2512.02793  (2025)

\bibitem{wu2025geometry}
Wu, H., Wu, D., He, T., Guo, J., Ye, Y., Duan, Y., Bian, J.: {Geometry Forcing:
  Marrying Video Diffusion and 3D Representation for Consistent World
  Modeling}. arXiv preprint arXiv:2507.07982  (2025)

\bibitem{wu2025cat4d}
Wu, R., Gao, R., Poole, B., Trevithick, A., Zheng, C., Barron, J.T., Holynski,
  A.: {Cat4D: Create Anything in 4D with Multi-View Video Diffusion Models}.
  In: Proceedings of the IEEE/CVF Conference on Computer Vision and Pattern
  Recognition (CVPR). pp. 26057--26068 (2025)

\bibitem{wu2025video}
Wu, T., Yang, S., Po, R., Xu, Y., Liu, Z., Lin, D., Wetzstein, G.: {Video World
  Models with Long-term Spatial Memory}. arXiv preprint arXiv:2506.05284
  (2025)

\bibitem{xiao2025worldmem}
Xiao, Z., Lan, Y., Zhou, Y., Ouyang, W., Yang, S., Zeng, Y., Pan, X.:
  {WorldMem: Long-Term Consistent World Simulation with Memory}. arXiv preprint
  arXiv:2504.12369  (2025)

\bibitem{xie2024carve3d}
Xie, D., Li, J., Tan, H., Sun, X., Shu, Z., Zhou, Y., Bi, S., Pirk, S.,
  Kaufman, A.E.: {Carve3D: Improving Multi-view Reconstruction Consistency for
  Diffusion Models with RL Finetuning}. In: Proceedings of the IEEE/CVF
  Conference on Computer Vision and Pattern Recognition (CVPR). pp. 6369--6379
  (2024),
  \url{https://openaccess.thecvf.com/content/CVPR2024/papers/Xie_Carve3D_Improving_Multi-view_Reconstruction_Consistency_for_Diffusion_Models_with_RL_CVPR_2024_paper.pdf}

\bibitem{xie2024moving}
Xie, J., Yang, C., Xie, W., Zisserman, A.: Moving object segmentation: All you
  need is sam (and flow). In: Proceedings of the Asian conference on computer
  vision. pp. 162--178 (2024)

\bibitem{xu2025depthsplat}
Xu, H., Peng, S., Wang, F., Blum, H., Barath, D., Geiger, A., Pollefeys, M.:
  {DepthSplat: Connecting Gaussian Splatting and Depth}. In: Proceedings of the
  IEEE/CVF Conference on Computer Vision and Pattern Recognition (CVPR). pp.
  16453--16463 (2025)

\bibitem{xu2024grm}
Xu, Y., et~al.: {GRM: Large Gaussian Reconstruction Model for Efficient 3D
  Reconstruction from Sparse Views}. arXiv preprint arXiv:2403.14621  (2024),
  \url{https://arxiv.org/abs/2403.14621}

\bibitem{xue2025dancegrpo}
Xue, Z., Wu, J., Gao, Y., Kong, F., Zhu, L., Chen, M., Liu, Z., Liu, W., Guo,
  Q., Huang, W., et~al.: Dancegrpo: Unleashing grpo on visual generation. arXiv
  preprint arXiv:2505.07818  (2025)

\bibitem{yang2019pointflow}
Yang, G., Huang, X., Hao, Z., Liu, M.Y., Belongie, S., Hariharan, B.:
  {PointFlow: 3D Point Cloud Generation with Continuous Normalizing Flows}. In:
  Proceedings of the IEEE/CVF International Conference on Computer Vision
  (ICCV). pp. 4541--4550 (2019)

\bibitem{yang2024cogvideox}
Yang, Z., Teng, J., Zheng, W., Ding, M., Huang, S., Xu, J., Yang, Y., Hong, W.,
  Zhang, X., Feng, G., et~al.: Cogvideox: Text-to-video diffusion models with
  an expert transformer. arXiv preprint arXiv:2408.06072  (2024)

\bibitem{yu2024representation}
Yu, S., Kwak, S., Jang, H., Jeong, J., Huang, J., Shin, J., Xie, S.:
  {Representation Alignment for Generation: Training Diffusion Transformers is
  Easier Than You Think}. arXiv preprint arXiv:2410.06940  (2024)

\bibitem{yu2024viewcrafter}
Yu, W., Xing, J., Yuan, L., Hu, W., Li, X., Huang, Z., Gao, X., Wong, T.T.,
  Shan, Y., Tian, Y.: {ViewCrafter: Taming Video Diffusion Models for
  High-Fidelity Novel View Synthesis}. arXiv preprint arXiv:2409.02048  (2024)

\bibitem{yuan2024spindiffusion}
Yuan, H., Chen, Z., Ji, K., Gu, Q.: {Self-Play Fine-Tuning of Diffusion Models
  for Text-to-Image Generation}. In: Advances in Neural Information Processing
  Systems (NeurIPS) (2024),
  \url{https://proceedings.neurips.cc/paper_files/paper/2024/file/860c1c657deafe09f64c013c2888bd7b-Paper-Conference.pdf}

\bibitem{zhang2025frame}
Zhang, L., Cai, S., Li, M., Wetzstein, G., Agrawala, M.: Frame context packing
  and drift prevention in next-frame-prediction video diffusion models. arXiv
  preprint arXiv:2504.12626  (2025)

\bibitem{zhang2018unreasonable}
Zhang, R., Isola, P., Efros, A.A., Shechtman, E., Wang, O.: The unreasonable
  effectiveness of deep features as a perceptual metric. In: Proceedings of the
  IEEE conference on computer vision and pattern recognition. pp. 586--595
  (2018)

\bibitem{zhang2025videorepa}
Zhang, X., Liao, J., Zhang, S., Meng, F., Wan, X., Yan, J., Cheng, Y.:
  Videorepa: Learning physics for video generation through relational alignment
  with foundation models. arXiv preprint arXiv:2505.23656  (2025)

\bibitem{zheng2025vbench2}
Zheng, D., Huang, Z., Liu, H., Zou, K., He, Y., Zhang, F., Zhang, Y., He, J.,
  Zheng, W.S., Qiao, Y., Liu, Z.: {VBench-2.0}: Advancing video generation
  benchmark suite for intrinsic faithfulness. arXiv preprint arXiv:2503.21755
  (2025)

\bibitem{zheng2024cami2v}
Zheng, G., et~al.: {CamI2V: Camera-Controlled Image-to-Video Diffusion Model}.
  arXiv preprint arXiv:2410.15957  (2024),
  \url{https://arxiv.org/abs/2410.15957}

\end{thebibliography}

\newcommand{\beginsupplement}{%
    \setcounter{table}{0}
    \renewcommand{\thetable}{S\arabic{table}}%
    \renewcommand{\theHtable}{S.\arabic{table}}%
    \setcounter{figure}{0}
    \renewcommand{\thefigure}{S\arabic{figure}}%
    \renewcommand{\theHfigure}{S.\arabic{figure}}%
    \setcounter{section}{0}
    \renewcommand{\thesection}{S\arabic{section}}%
    \renewcommand{\theHsection}{S.\arabic{section}}%
    \renewcommand{\theHsubsection}{S.\arabic{section}.\arabic{subsection}}%
    \setcounter{equation}{0}
    \renewcommand{\theequation}{S\arabic{equation}}%
    \renewcommand{\theHequation}{S.\arabic{equation}}%
    \renewcommand{\thepage}{S\arabic{page}}%
}
\clearpage
\setcounter{page}{1}
\beginsupplement

\section{Supplementary Website}
Please refer to the supplementary website for high-resolution video results. It includes video versions of all examples from the main paper, along with additional qualitative comparisons and failure cases.

\section{Further Ablations}
In this section, we present two additional ablation studies. First, we analyze the impact of substituting the foundational geometric and motion models used in our reward pipeline. Second, we isolate and ablate the individual components of the proposed composite reward.

\subsection{Base Model Robustness}
To demonstrate that our reward formulation is robust and not overly sensitive to the exact choice of the underlying estimators, we perform an ablation where we swap the base models used for optical flow and depth prediction. Specifically, we replace WAFT~\cite{wang2025waft} with RAFT~\cite{teed2020raft}, and Depth Anything 3~\cite{lin2025depth} with Pi3X~\cite{wang2025pi}. We evaluate these configurations using the identical protocol described in the main paper. We additionally report the average optical flow magnitude (in pixels) as a proxy for the dynamic degree of the video (see \cref{sec:dynamic_degree} for a detailed discussion on this metric).

The results, shown in \cref{tab:sup_model}, validate the hypothesis that the reward's effectiveness is largely independent of the specific base models. Exchanging WAFT with RAFT yields comparable performance: while MEt3R degrades slightly, the videos exhibit a marginally higher dynamic degree, and the Sampson error is slightly reduced. A more pronounced shift occurs when replacing Depth Anything 3 with Pi3X; the geometric consistency metrics improve further, but the overall motion magnitude drops. At these extreme levels of consistency, the model begins to exhibit a direct trade-off between strict structural rigidity and natural motion dynamics.

\begin{table}[t]
    \centering
    \caption{\textbf{Base model ablation.} Impact of replacing components of our reward during training.}
    \label{tab:sup_model}
    \small
    \setlength{\tabcolsep}{1pt}
    \begin{tabular}{lcccc}
        \toprule
        \textbf{Configuration} & \textbf{MEt3R} $\downarrow$ & \textbf{Sampson} $\downarrow$ & \textbf{Flow} & \textbf{Aes.~Qual.} $\uparrow$ \\
        \midrule
        Base Model  & 8.212 & 5.453 & 6.943 & 0.582 \\
        w/ RAFT  & 4.823  & 2.623  & 2.706  &  0.591 \\
        w/ Pi3X & 3.423 & 1.609 & 1.804 & 0.598 \\
        Ours & 4.594 & 2.709  & 2.579  & 0.589 \\
        \bottomrule
    \end{tabular}
\end{table}

\subsection{Reward Component Ablation}
\begin{table}[t]
    \centering
    \caption{\textbf{Reward component ablation.} Impact of removing components of the reward during training.}
    \label{tab:sup_partial}
    \small
    \setlength{\tabcolsep}{1pt}
    \begin{tabular}{lcccc}
        \toprule
        \textbf{Configuration} & \textbf{MEt3R} $\downarrow$ & \textbf{Sampson} $\downarrow$ & \textbf{Flow} & \textbf{Aes.~Qual.} $\uparrow$ \\
        \midrule
        Base Model  & 8.212 & 5.453 & 6.493 & 0.582 \\
        w/o Dino  & 6.791  & 3.829  & 5.761  &  0.578 \\
        w/o Geo & 4.283 & 2.345 & 1.892 & 0.601 \\
        Full & 4.594 & 2.709  & 2.579  & 0.589 \\
        \bottomrule
    \end{tabular}
\end{table}
Our composite reward comprises two complementary signals: (1) a structural geometric score (\textit{Geo}) that relies on depth and optical flow to penalize deviations from rigid, camera-induced background motion, and (2) a semantic consistency score (\textit{DINO}) that ensures all parts of the scene preserve identity along flow trajectories. We isolate their respective impacts in \cref{tab:sup_partial}. 

We observe that relying solely on the structural geometric score (w/o DINO) improves basic consistency metrics while largely maintaining the dynamic degree of the video. This behavior is expected, as the rigid geometric reward acts as a soft weight and does not strongly penalize independently moving foreground objects. Conversely, using only the semantic score (w/o Geo) evaluates both rigid and dynamic regions indiscriminately. While this seemingly leads to superior consistency scores compared to the full reward, it substantially reduces the dynamic degree of the generated videos. The full composite reward strikes the optimal balance, enforcing background stability without stifling natural object motion.

\section{Adapting Metrics for Dynamic Scenes}
\begin{table*}[t]
\centering
\caption{\textbf{Quantitative comparison of geometric consistency}. We compare against Wan2.1 and VideoRepa across a $2 \times 2$ matrix of motion regimes: Static vs. Dynamic scenes, and Simple vs. Complex descriptions. Arrows indicate whether higher ($\uparrow$) or lower ($\downarrow$) is better. In contrast to the main consistency table, these metrics use the dynamic adaptations, and Gemini measures dynamic degree.}
\label{tab:sup_main_results}
\begin{tabular}{ll ccc c ccc}
\toprule
& & \multicolumn{3}{c}{\textbf{Simple}} & & \multicolumn{3}{c}{\textbf{Complex}} \\
\cmidrule{3-5} \cmidrule{7-9}
\textbf{Type} & \textbf{Method} & MEt3R~$\downarrow$ & Sampson~$\downarrow$ & Gemini~$\uparrow$ & & MEt3R~$\downarrow$ & Sampson~$\downarrow$ & Gemini~$\uparrow$ \\
\midrule
\multirow{3}{*}{\textbf{Static}} 
& Wan2.1    & 7.340 & 1.693 & 6.317 & & 6.420 & 2.048 & 5.599 \\
& VideoRepa & 5.320 & \textbf{0.859} & 5.793 & & 6.740 & 1.818 & 5.552 \\
& Ours      & \textbf{4.510} & 0.892 & \textbf{6.499} & & \textbf{3.110} & \textbf{0.972} & \textbf{5.804} \\
\midrule
\multirow{3}{*}{\textbf{Dynamic}} 
& Wan2.1    & 13.088 & 2.366 & 6.012 & & 12.797 & 4.659 & 6.132 \\
& VideoRepa & 8.695 & \textbf{1.274} & 5.502 & & 10.923 & 4.018 & 5.788 \\
& Ours      & \textbf{7.568} & 1.999 & \textbf{6.545} & & \textbf{7.543} & \textbf{3.035} & \textbf{6.453} \\
\bottomrule
\end{tabular}
\end{table*}
While we report established multi-view consistency metrics (MEt3R~\cite{asim2025met3r} and Sampson Error~\cite{sampson1982fitting}) in the main paper, these metrics are traditionally formulated under a strict static-scene assumption. In this section, we detail how we adapted these metrics to reliably evaluate dynamic videos.

The authors of MEt3R note the possibility of using an optical flow model to compare the warped pixels to the real video. They note that this is not as precise as their final MEt3R model. However, their default MASt3R~\cite{leroy2024grounding} backend does not support dynamic scenes. In contrast, the RAFT version combined with LPIPS~\cite{zhang2018unreasonable} can score dynamic scenes. In this section's experiments, we use this configuration.

To make the Sampson error compatible with dynamic scenes, we compute a moving object mask using FlowSAM~\cite{xie2024moving}. Next, we extract and match SIFT~\cite{lowe2004distinctive} features, retaining only the matches where both keypoints fall entirely outside the dynamic mask. %
Then, we proceed as usual with the computation of the Sampson error.

\cref{tab:sup_main_results} shows the results we obtain with these metrics on the same data as presented in the main paper. Our method strongly improves consistency across all categories. It also performs better on average than VideoRepa, which substantially degrades visual quality.

\section{Further Analysis}
Here, we discuss several nuanced aspects of optimizing for geometric consistency. We explore the relationship between consistency and the dynamic degree of generated videos, clarify what constitutes a successful consistency improvement, visualize the reward signal, and explicitly discuss observed failure modes.

\subsection{Dynamic Degree and Motion Perception} \label{sec:dynamic_degree}
Qualitatively, one might observe that videos aligned for strict geometric consistency feel less dynamic. Quantitatively, as reported in \cref{tab:sup_model}, the average optical flow magnitude is reduced from 6.9 to 2.6 pixels. However, interpreting these raw flow numbers in isolation is misleading. Standard unaligned video generators frequently produce high optical flow values not due to natural, intentional movement, but because of severe artifacts: texture drift, background wobble, and object deformation. 

Furthermore, off-the-shelf flow estimators like RAFT~\cite{teed2020raft} are prone to errors when presented with the highly inconsistent, unrealistic frames typical of unaligned models. Consequently, raw flow magnitude confounds genuine scene dynamics with artifactual pixel shifting. To automatically and reliably separate wanted movement from unwanted geometric noise, we rely on the visual reasoning capabilities of Gemini 3.0-Flash. By prompting the VLM to rate the dynamic degree while explicitly discounting structural artifacts, we obtain a much more accurate representation of true motion preservation.

The results are shown in \cref{tab:sup_main_results}. Our method has the best dynamics score after accounting for distortions and artifacts. While VideoRepa also produces more consistent generations, it produces far less dynamic content than both Wan and our method.

\subsection{Defining Geometric Improvement}
The overarching goal of GeoFlow is to improve spatial and temporal coherence. However, defining improvement in the context of generative alignment is complex. Ideally, RL fine-tuning should act as an orthogonal enhancement: we hope to eliminate geometric artifacts without altering the original scene layout, the subject's identity, the prompt alignment, or the intended camera trajectory. With flow-matching models~\cite{lipman2022flow} the only variation stems from the initial noise, so it determines the scene structure and movements. This means that we would hope that sampling a baseline model and the aligned model with the same prompt and initial noise generate the same scene just with improved consistency.

Measuring this multi-dimensional preservation exactly is extremely challenging. However, our qualitative observations confirm that GeoFlow successfully optimizes the inconsistent portions of a generated trajectory while mostly anchoring the overall semantic and aesthetic layout. Please refer to the side-by-side trajectory comparisons on our supplementary website, which demonstrate how object permanence is maintained while background morphing is eliminated. We describe in the captions under the videos how scene layouts and camera movements are kept. In cases where there is little hope of improvement around the initial denoising trajectory it also happens that a completely new but more consistent scene is generated.

\subsection{Reward Visualization}
\begin{figure*}[th]
    \centering
    \includegraphics[width=1.0\linewidth]{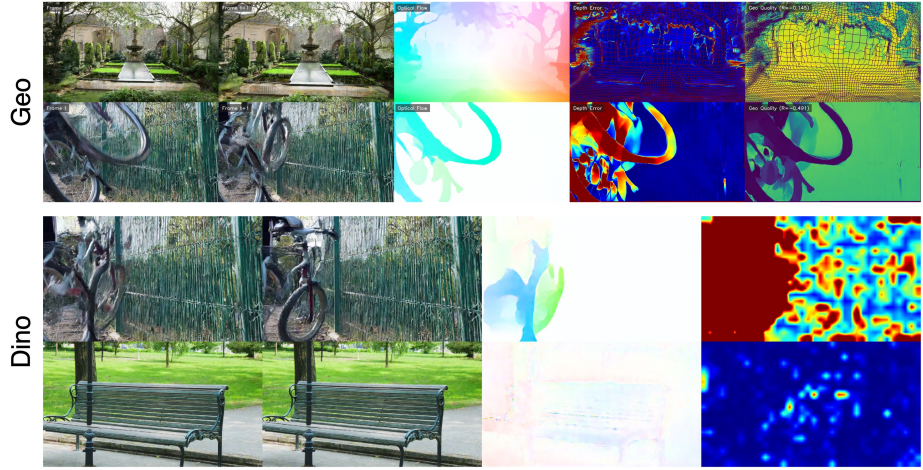}
    \caption{\textbf{Reward visualization.} The top two rows show how the Geo reward component is computed, and the bottom row shows the influence of the DINO component. The Geo and DINO rewards complement each other: for the bicycle example, the Geo reward penalizes the inconsistent background, while the DINO component penalizes the morphing bicycle.}
    \label{fig:sup_reward}
\end{figure*}
To better illustrate the learning signal, \cref{fig:sup_reward} visualizes the individual reward components across sample input sequences. The first row demonstrates how the geometry (geo) reward penalizes structural inconsistencies. Notably, the significant depth error on the left-side tree is mitigated by the high local motion in that region, yielding a moderate geo score of -14.5. The bicycle example exhibits a similar dynamic: the bike itself has a larger depth error but high motion, whereas the static background has a smaller depth error but zero motion. The geo reward heavily penalizes the static background inconsistencies, resulting in a substantially worse score of -49.1. The subsequent row details the DINO reward, which complements the geometric evaluation by specifically targeting semantic degradation. It effectively captures the morphing artifacts on the bicycle, assigning a penalty of -33.8. Finally, a temporally consistent image pair incurs minimal DINO penalty, achieving a near-optimal score of -3.8.

\subsection{Failure Cases}
While our reward is designed to handle dynamic scenes gracefully, it exhibits limitations in scenarios involving severe topological changes. For instance, when an object undergoes massive occlusion, or when a structural barrier is broken (e.g., opening a closed drawer to reveal new contents), the fundamental assumption of pixel persistence is violated. In such cases, the optical flow estimators fail to find valid correspondences, leading to noisy reward signals. Conversely, some rapid changes cannot be penalized if they happen faster than the sampling interval of the reward.

Additionally, the RL optimization process is inherently bounded by the capabilities of the base model. Our method relies on the model generating at least partially coherent structures early in the denoising process. If the base model's generations for a specific, out-of-distribution prompt are fundamentally chaotic or lack a basic 3D foundation, the reward signal may be too sparse to guide the policy toward a geometrically consistent state.

Furthermore, we assume that the scene can be split into static and dynamic parts. If parts of the scene lie near this boundary, e.g., the waving tent on the website, the reward may remove this dynamic component even though it may be acceptable.

Finally, while the reward improves consistency in several dimensions, it mainly serves as an alignment method for an existing model. Because the model remains close to its original distribution, it is not possible to eliminate all inconsistencies; in some generations, consistency is only partially improved.

Please view the corresponding section of the website to see what these failure cases look like in video.

\section{Experiment Details}
In this section, we provide the comprehensive hyperparameters and configurations necessary to replicate our results. 

\subsection{Dataset Construction}

To train and evaluate our geometry-aware policy, we construct a curated, motion-annotated prompt dataset. We pool video captions from OpenVid-1M~\cite{nan2024openvid} and DL3DV~\cite{ling2024dl3dv}, oversampling candidate prompts at $3\times$ our target count. Using keyword heuristics, we assign each prompt to a specific spatial dynamic regime (e.g., \textsc{Static Scene}, \textsc{Camera Around}) to ensure a balanced distribution of motion behaviors.

\paragraph{LLM-assisted Standardization.} 
Raw captions frequently contain noisy or stylized language that confounds motion conditioning. To distill prompts to pure geometric intent, we use Gemini 3.0-Flash~\cite{team2023gemini} to rewrite all candidates. We enforce strict linguistic constraints: prompts must be exactly 1--2 sentences and 8--35 words, use simple present tense, and contain explicit motion verbs. For evaluation, we allow more complex captions with over 100 words.

\paragraph{Filtering and Balancing.} 
Following standardization, we automatically reject prompts lacking clear noun anchors, unambiguous motion, or those violating vocabulary rules. To prevent semantic collapse and ensure subject diversity, we deduplicate the remaining pool by clustering sentence embeddings within each regime. Finally, we randomly sample the exact target counts required per regime while enforcing source-domain quotas. A final automated quality check ensures all selected prompts possess grammatical correctness and definitive motion labels, yielding a pristine dataset for stable RL optimization.

\subsection{Training Details}
We apply our geometry-consistency reward to fine-tune the \texttt{Wan2.1-T2V-1.3B} base model~\cite{wan2025wan} using Flow-GRPO. We train at a spatial resolution of $480 \times 832$, generating $33$ frames per video. Training is executed across 16 NVIDIA H100 GPUs using \texttt{bf16} precision. 

During the online rollout phase, each GPU processes 4 prompts, sampling $G=4$ candidate videos per prompt with a classifier-free guidance scale of $4.5$. Group advantages are normalized using per-prompt statistical tracking rather than global standardization. To ensure stable policy optimization, we synchronize the initial latent noise $\boldsymbol{\epsilon}$ across all $G$ candidates. 

During the optimization phase, we use a per-GPU micro-batch size of 2, yielding an effective global batch size of 128 (2 batches per epoch). We update the LoRA~\cite{hu2022lora} weights using the standard AdamW~\cite{kingma2014adam} optimizer (without 8-bit quantization). We set the learning rate to $1 \times 10^{-4}$, weight decay to $1 \times 10^{-4}$, $\beta_1 = 0.9$, $\beta_2 = 0.999$, and $\epsilon = 1 \times 10^{-8}$. To stabilize training gradients, we apply gradient clipping with a maximum norm of $1.0$.

The policy update is constrained by a clipping range of $1 \times 10^{-3}$ and regularized by a KL divergence penalty coefficient of $\beta = 0.004$. We additionally apply Exponential Moving Average (EMA) to the LoRA weights. Utilizing our full-resolution truncated backpropagation strategy, we generate candidate videos over a full sequence of $40$ SDE steps, but only track gradients and evaluate expectations over the first $20$ steps of the reverse trajectory. 

\subsection{Reward Computation Details}
The aggregate consistency reward is calculated for consecutive frame pairs and then averaged across the sequence. We equally weight the structural and semantic components, setting $\lambda = 0.5$ such that $R = 0.5 \cdot R_{\mathrm{geo}} + 0.5 \cdot R_{\mathrm{dino}}$. 
We set $\epsilon$ to 1.5px.
For our primary configuration, we extract metric depth and camera intrinsics/extrinsics using Depth Anything 3 Large (v1.1)~\cite{lin2025depth}, predict optical flow using WAFT~\cite{wang2025waft}, and extract semantic patch features using DINOv2-base~\cite{oquab2023dinov2}. Furthermore, we optionally apply confidence gating, weighting both $R_{\mathrm{geo}}$ and $R_{\mathrm{dino}}$ by the per-pixel confidence maps generated by Depth Anything 3 to mitigate the impact of unreliable depth estimates.

\subsection{Evaluation Details}
We evaluate performance on a curated dataset of 110 unique prompts sourced from OpenVid-1M~\cite{nan2024openvid} and DL3DV~\cite{ling2024dl3dv}, augmented via Gemini 3.0-Flash to encompass both static/dynamic and simple/complex categorizations. During evaluation, models generate videos using $50$ denoising steps and a guidance scale of $4.5$. Each prompt is sampled with 5 different initial random seeds to minimize the influence of initial latent structure, yielding 550 evaluated videos per configuration. 

For visual quality preservation, we utilize the standardized VBench 1.0~\cite{huang2024vbench} protocol, specifically reporting on Background Consistency, Aesthetic Quality, Temporal Flickering, and Motion Smoothness.

Next, we provide the prompts used to ask Gemini to evaluate consistency and dynamic degree.

\begin{tcolorbox}[colback=gray!5!white,colframe=gray!75!black,title=Consistency Evaluation Prompt]
You are a video quality evaluator. You are given \{n\_frames\} frames sampled \
evenly from a generated video. The original prompt for the video was: \
"\{prompt\}"

Rate the **temporal and visual consistency** of this video on a scale from \
0.0 to 10.0 (one decimal place), where:
  0.0 = Completely broken (random noise, no coherent content)
  2.0 = Very inconsistent (heavy flickering, morphing objects)
  5.0 = Moderate consistency (some flickering or object deformation)
  8.0 = Good consistency (minor artifacts, mostly smooth)
  10.0 = Perfectly consistent (smooth motion, stable objects, no artifacts)

Consider:
- Do objects maintain their shape and identity across frames?
- Is the motion smooth and physically plausible?
- Are there any sudden jumps, flickering, or morphing artifacts?
- Is the lighting consistent across frames?

Respond with ONLY a single number between 0.0 and 10.0 (one decimal place). \
Nothing else.
\end{tcolorbox}

\begin{tcolorbox}[colback=gray!5!white,colframe=gray!75!black,title=Dynamic Evaluation Prompt]
You are a video quality evaluator specializing in motion quality. You are \
given \{n\_frames\} frames sampled evenly from a generated video. The original \
prompt for the video was: "\{prompt\}"

Rate the **motion quality** of this video on a scale from 0.0 to 10.0 \
(one decimal place). You should reward videos that have meaningful motion \
AND maintain visual quality during that motion. Heavily penalize any \
motion-induced errors or artifacts.

Scoring rubric:
  0.0 = Severe motion-induced distortions (objects warp, deform, or dissolve \
during movement; faces morph; limbs distort; heavy flickering)
  2.0 = Significant motion artifacts (noticeable warping or morphing during \
movement, objects losing shape, inconsistent geometry)
  4.0 = Some motion artifacts present (occasional distortion during fast \
movement, slight object deformation, minor flickering)
  5.0 = Mostly static with no artifacts, OR moderate motion with minor issues
  7.0 = Clear motion present with good consistency (objects maintain shape \
and identity while moving, smooth transitions)
  9.0 = Highly dynamic with excellent consistency (fast or complex motion \
with no visible artifacts, stable objects, physically plausible movement)
  10.0 = Perfect---significant, complex motion that is completely clean and \
physically plausible

Key criteria (in order of importance):
1. Do moving objects maintain their shape, structure, and identity? \
(Heavily penalize warping, morphing, melting, or dissolving)
2. Are there distortions at motion boundaries? (ghosting, smearing, tearing)
3. Is the motion physically plausible? (no impossible deformations)
4. How much meaningful motion is present? (more motion = higher ceiling, \
but only if clean)

A static video with no artifacts should score around 5.0. \
A dynamic video with artifacts should score BELOW 5.0. \
A dynamic video without artifacts should score ABOVE 5.0.

Respond with ONLY a single number between 0.0 and 10.0 (one decimal place). \
Nothing else.
\end{tcolorbox}

\immediate\closein\imgstream

\end{document}